\documentclass[10pt,twocolumn,letterpaper]{article}

\usepackage[pagenumbers]{cvpr} 
\usepackage[table]{xcolor}
\usepackage{subcaption}
\usepackage{graphicx}
\usepackage{multirow}
\definecolor{cvprblue}{rgb}{0.21,0.49,0.74}
\usepackage[pagebackref,breaklinks,colorlinks,allcolors=cvprblue]{hyperref}

\def\paperID{*****} 
\def\confName{CVPR}
\def\confYear{2026}

\title{ImVideoEdit: Image-learning Video Editing via 2D Spatial Difference Attention Blocks}
\author{
    \textbf{Jiayang Xu\thanks{Equal Contribution.}\ , \ Fan Zhuo$^{*}$\ , \ Majun Zhang$^{*}$\ , \ Changhao Pan\ ,} \\
    \ \textbf{Zehan Wang\thanks{Project Leader.}\ , \ Siyu Chen\ , \ Xiaoda Yang\ , \ Tao Jin\ , \ Zhou Zhao\thanks{Corresponding Author.}}
    \\ 
    Zhejiang University
    \\
    {\tt\small jiayangxu@zju.edu.cn} \quad {\tt\small zhaozhou@zju.edu.cn}
}
\begin{document}

\maketitle
\begin{figure*}[t]
  \centering
  \includegraphics[width=\textwidth]{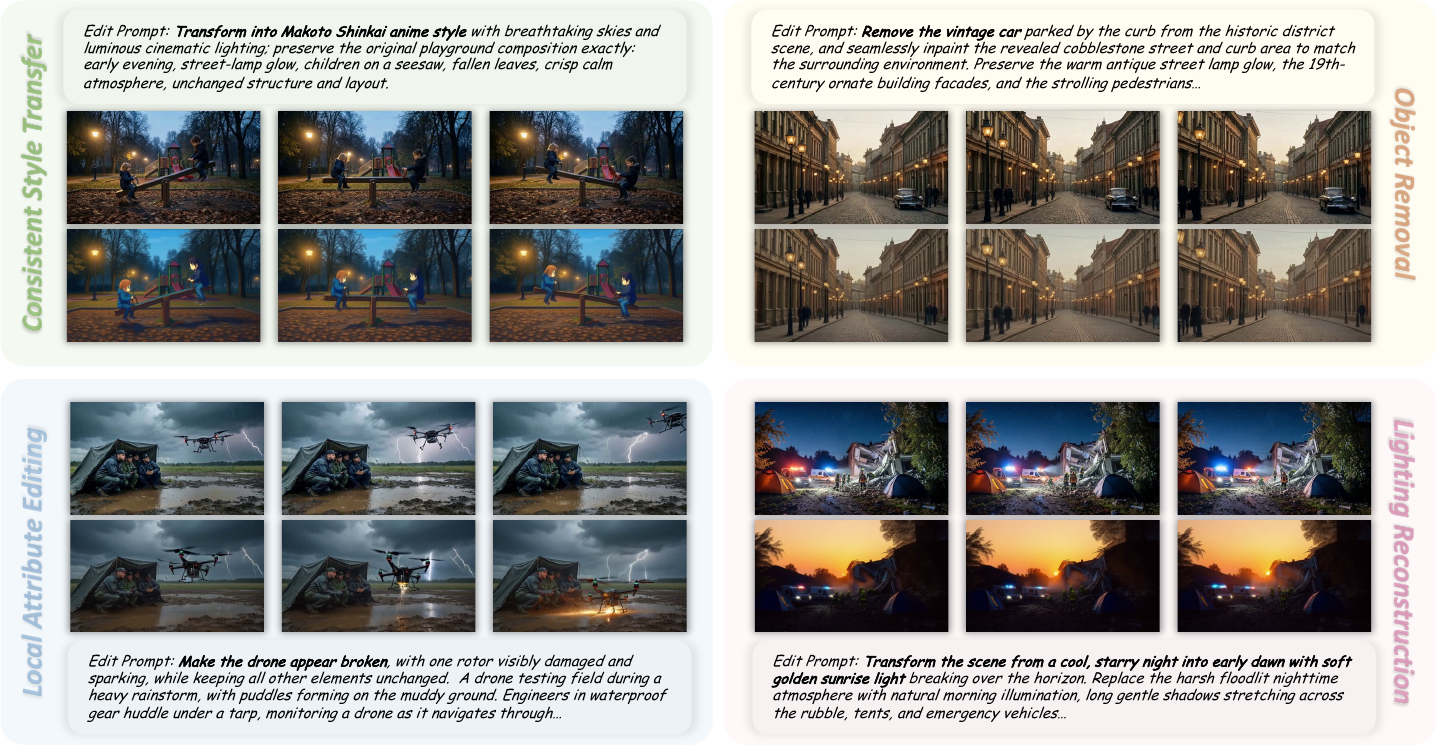}
  \caption{Illustration of four of \textit{ImVideoEdit}’s basic editing task. Zoom in for best viewing.}
  \label{fig:teaser}
\end{figure*}


\begin{abstract}

Current video editing models often rely on expensive paired video data, which limits their practical scalability. In essence, most video editing tasks can be formulated as a decoupled spatiotemporal process, where the temporal dynamics of the pretrained model are preserved while spatial content is selectively and precisely modified. Based on this insight, we propose \textit{ImVideoEdit}, an efficient framework that learns video editing capabilities entirely from image pairs.  By freezing the pre-trained 3D attention modules and treating images as single-frame videos, we decouple the 2D spatial learning process to help preserve the original temporal dynamics. The core of our approach is a \textit{Predict-Update Spatial Difference Attention} module that progressively extracts and injects spatial differences. Rather than relying on rigid external masks, we incorporate a \textit{Text-Guided Dynamic Semantic Gating} mechanism for adaptive and implicit text-driven modifications. Despite training on only 13K image pairs for 5 epochs with exceptionally low computational overhead, \textit{ImVideoEdit} achieves editing fidelity and temporal consistency comparable to larger models trained on extensive video datasets.

\end{abstract}

\section{Introduction} \label{sec:Introduction}


Diffusion models, particularly 3D Diffusion Transformers (3D DiTs), have achieved revolutionary breakthroughs in video generation, as demonstrated by cutting-edge models like Seedance \cite{seedance2025seedance, gao2025seedance} and Veo \cite{veo3techreport2025}. However, generating high-quality videos is merely the first step. Real-world content creation demands not only superior generation but also robust editing capabilities that strike a balance between semantic manipulation and structural preservation. Specifically, this requires the ability to execute precise modifications on existing videos guided by textual prompts.


Despite significant progress in recent video editing, existing approaches face a fundamental dilemma when adapted to 3D DiTs. Direct fine-tuning or feature injection into the highly coupled 3D spatio-temporal attention modules severely disrupts the delicate motion priors of pre-trained models, typically culminating in background drift and temporal flickering. To circumvent this instability, current pipelines frequently resort to external foundation segmentation models or manually annotated masks. Yet, this reliance compromises the elegance of end-to-end, text-driven editing and proves fundamentally brittle when handling complex non-rigid deformations, thereby sacrificing the potential for zero-shot interaction.


Compounding these architectural challenges is a severe bottleneck in data acquisition. Constructing large-scale, diverse paired datasets—comprising source videos, target videos, and text instructions—imposes a prohibitive cost barrier. While scaling up massive amounts of data can undeniably yield models with strong generalization capabilities, the inherent complexity of dynamic video scenes makes acquiring and synthesizing such paired data prohibitively expensive and time-consuming. This inherent data scaling bottleneck significantly inflates the computational and temporal costs required to develop open-domain video editing models.

Returning to the essence of the video editing task, it fundamentally revolves around the precise reorganization and modification of the original video's features. Many video editing applications, such as style transfer and object addition, removal, or modification, primarily entail the reconstruction of spatial features while rigorously preserving the underlying temporal dynamics. Consequently, the heavily coupled spatiotemporal features inherent in video data are, to some extent, temporally redundant for training these predominantly spatial editing tasks. 
Fortunately, as demonstrated by recent pioneering works such as ViFeEdit \cite{yu2026vifeeditvideofreetunervideo}, image editing data, which naturally isolates and focuses exclusively on spatial feature transformations, has proven to be a highly effective surrogate to facilitate the training of video editing models.
Building upon this insight, we further propose \textit{\textbf{ImVideoEdit}}, an innovative method learning video editing from images via 2D spatial difference attention blocks.

In order to generate a high-quality dataset, we design a three-stage pipeline: scene-conditioned prompt construction, paired image synthesis, and data filtering. This process produces approximately 13K high-quality image pairs that provide dense supervision for learning video editing. The dataset encompasses a wide variety of scene compositions and editing tasks, offering diverse and robust training signals for spatial feature transformation.

Since our paradigm treats images as single-frame videos to learn 2D spatial feature reorganization, it is imperative to preserve the model's inherent spatiotemporal modeling capabilities. Following established practices, we completely freeze the backbone of the pre-trained video diffusion model (Wan2.1-T2V-1.3B). This strategy safely safeguards the robust spatiotemporal priors and motion dynamics encapsulated within its 3D self-attention mechanisms, which were learned during large-scale pre-training. Thus, the crux of the problem converges on a singular, critical challenge: \textit{\textbf{how to optimally extract and inject the 2D spatial features of the source video without temporal interference?}}

To address this, drawing inspiration from the predictive-corrective paradigms utilized in camera-control video generation \cite{yang2026confctrlenablingprecisecamera}, we introduce an innovative \textbf{Predict-Update Spatial Difference Attention Module}. This architecture decouples the spatial feature reconstruction into a progressive two-step process. First, the \textit{Predict} phase establishes a coarse-grained spatial structural alignment. Subsequently, the \textit{Update} phase precisely captures and fits the high-frequency spatial residual differences. This Predict-Update mechanism empowers \textit{ImVideoEdit} to achieve exceptionally high-fidelity extraction and editing of the source video's spatial features. 
Furthermore, since our spatial module precedes the native cross-attention layers and inherently lacks text awareness, we introduce \textbf{Text-Guided Dynamic Semantic Gating} to enable prompt-driven, precise semantic modulation.

In summary, our main contributions are multi-fold:
\begin{itemize}
    \item \textbf{Dataset Construction:} We provide a curated dataset of 13K image pairs that supports learning spatial transformations in video editing tasks, offering rich supervision without relying on full video sequences.

    \item \textbf{Video-Free Training Paradigm \& Architectural Evolution:} Moving beyond computationally prohibitive video-based training, we pioneer an efficient paradigm that learns video editing entirely from static images. To enable this, we propose the \textbf{Predict-Update Spatial Difference Attention} module. By seamlessly treating images as single-frame videos and establishing a spatial residual stream, it achieves coarse-to-fine spatial feature extraction while safeguarding the fragile 3D spatiotemporal priors.

    \item \textbf{Zero-Shot Text-Driven Semantic Modulation:} To facilitate fine-grained and prompt-faithful video editing, we introduce the \textbf{Text-Guided Dynamic Semantic Gating}. Without relying on external masks, this design provides strong text-driven guidance during spatial feature learning. 
    
    \item \textbf{State-of-the-Art Performance:} Extensive evaluations demonstrate the superiority of \textit{ImVideoEdit}. This proves that robust, fine-grained video editing can be accomplished with minimal computational overhead and data dependency.
\end{itemize}


\section{Related Work}
\subsection{Video Generation with Diffusion Models}

Early approaches to video generation, including GANs and RNN-based methods~\cite{vondrick2016generating, denton2018stochastic}, were limited by poor temporal coherence and low visual fidelity. The introduction of diffusion-based architectures, particularly 3D U-Net variants~\cite{ho2022video, singer2022make, guo2023animatediff}, significantly improved spatiotemporal modeling and enabled the generation of high-quality short video clips.
More recently, Diffusion Transformers (DiTs) and large-scale generative architectures~\cite{peebles2023scalable} have driven rapid advancements in video foundation models. Representative systems such as HunyuanVideo \cite{kong2024hunyuanvideo}, Cosmos \cite{agarwal2025cosmos}, Wan \cite{wan2025wan}, and Kling \cite{team2025kling} demonstrate strong capabilities in synthesizing high-resolution, and physically plausible videos. These models benefit from scaling model capacity, improved architecture design, and training on large-scale curated video datasets, leading to substantial gains in realism, motion consistency, and multimodal alignment.

\begin{figure*}[t] 
  \centering
  \includegraphics[width=\textwidth]{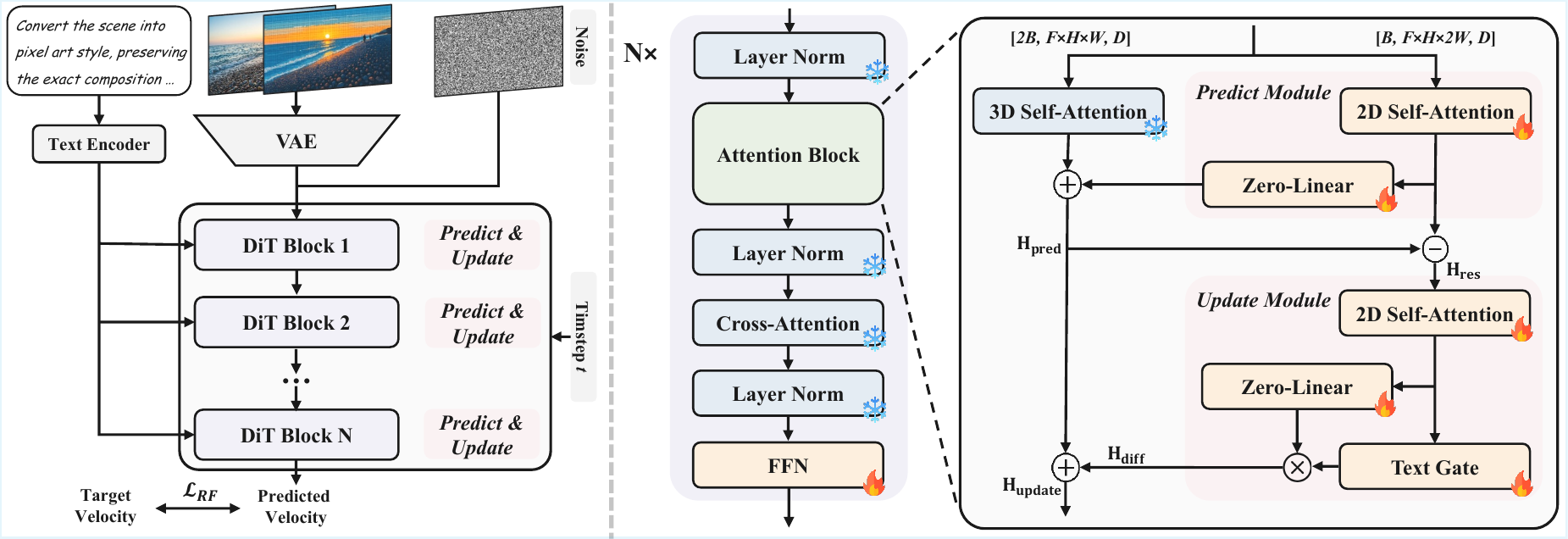} 

  \caption{Overview of \textit{ImVideoEdit}. Left: The overall pipeline processes latents from \textit{single image} through a frozen 3D DiT, featuring a Predict-Update module parallel to each attention block. Right: Detailed design of the \textit{Predict \& Update} Module. The frozen 3D self-attention safeguards spatiotemporal priors, while the parallel 2D branch extracts spatial features from the reference latent. The Predict Module generates a coarse alignment ($H_{pred}$) via a Zero-Linear layer. The spatial difference ($H_{res}$) is then processed by the Update Module and a Text Gate, which implicitly modulates the editing intensity to yield $H_{diff}$. The final output ($H_{update}$) achieves precise text-driven spatial reconstruction. During training stage, the numbers of frame $F$ maintains $1$.}
  \label{fig:model}
\end{figure*}

\subsection{Video Editing}
Early diffusion-based video editing paradigms primarily adapted Text-to-Image (T2I) models for the video domain, employing strategies such as cross-attention map injection or deterministic DDIM inversion ~\cite{qi2023fatezero, tang2023any}. However, directly applying these T2I inversion techniques to native Text-to-Video (T2V) architectures often introduces severe color flickering and structural distortions due to tightly coupled spatio-temporal representations.  To overcome the flickering and structural distortions inherent in early inversion-based methods, recent works have shifted towards the end-to-end training of native video generative models. Approaches such as ~\cite{tan2025omni, wei2025univideo, yu2025veggie, jiang2025vace} integrate Multimodal Large Language Models (MLLMs) with Diffusion Transformers to unify diverse editing tasks into a single architecture. Moreover, to overcome the data scarcity bottleneck in training end-to-end video editing models, methods like ~\cite{he2025openve, lin2026kiwi, bai2025scaling} have proposed large-scale synthetic video generation pipelines. However, end-to-end training on such massive video datasets inevitably incurs substantial computational overhead and high data generation costs. To circumvent this heavy reliance on exhaustive video-level optimization, we propose a novel approach that effectively achieves temporally coherent video editing by training exclusively on image editing data.

\subsection{Attention Control and Spatial Adapters}

Achieving fine-grained, high-fidelity synthesis in generative models relies heavily on manipulating internal representations, predominantly through attention control and spatial adapters. Building upon the foundational cross-attention manipulation of Prompt-to-Prompt ~\cite{hertz2022prompt}, recent advancements in attention control enable training-free semantic editing while strictly preserving structural integrity. Techniques leveraging localized and relative guidance ~\cite{xiao2025fastcomposer, zhang2025group}, alongside region-selective denoising ~\cite{yin2025consistedit, qin2025spotedit}, effectively mitigate identity blending and anchor background fidelity during complex foreground transformations.  Complementary to these internal mechanisms, spatial adapters ~\cite{mou2024t2i, tan2025ominicontrol} provide parameter-efficient paradigms to inject external structural priors into pre-trained latents. These integrated priors encompass a wide spectrum of conditioning signals, ranging from dense visual maps to sparse grounding tokens ~\cite{li2023gligen} and relational scene graphs ~\cite{shen2024sg}, which eventually culminate in unified frameworks for precise dual-control ~\cite{ nguyen2025universal}. While these methodologies have established exceptional spatial layout guidance and high-fidelity semantic editing in the image domain, extending such granular control to videos remains notoriously difficult due to the lack of robust temporal consistency. Motivated by the rich spatial and semantic priors encapsulated in these image-based frameworks, our work proposes to leverage image editing data to train a robust video editing model, effectively transferring sophisticated frame-level control to the temporal domain.

\section{Dataset}

\begin{figure}[t] 
  \centering
  \includegraphics[width=0.9\linewidth]{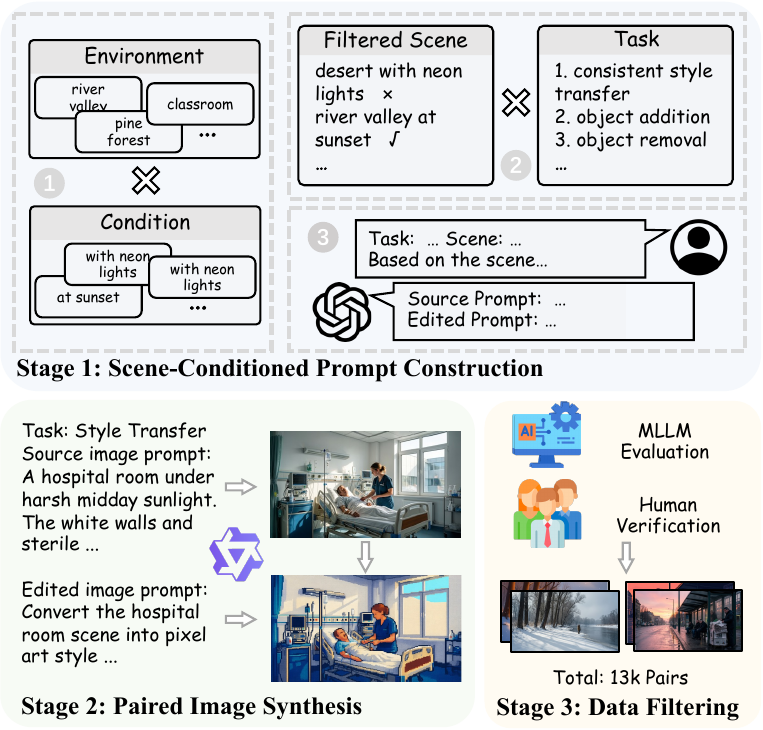} 
  \caption{Overview of the dataset construction pipeline. }
  \label{fig:pipeline}
\end{figure}

To enable instruction-driven video editing under image-level supervision, we construct a paired image dataset that explicitly captures diverse editing operations together with their resulting visual outcomes. As illustrated in Fig.~\ref{fig:pipeline}, our data pipeline consists of three stages: scene-conditioned prompt construction, paired image synthesis, and data filtering.

\noindent\textbf{Scene-Conditioned Prompt Construction.}
We first define a set of base environments (or entities) 
\(
\mathcal{E} = \{e_1, e_2, \dots, e_n\}
\)
, together with a set of compositional conditions 
\(
\mathcal{C} = \{c_1, c_2, \dots, c_m\}
\)
. For each environment \( e \in \mathcal{E} \), we randomly sample conditions from \( \mathcal{C} \) and combine them to form a base scene description. This compositional construction enables scalable generation of diverse scenes while maintaining controllability.
However, arbitrary combinations often yield semantically implausible or visually incoherent scenes. To mitigate this, we leverage Gemini 3.1 Pro \cite{google2025gemini} for semantic validation, filtering out descriptions that violate physical principles, defy commonsense, or lack clear visual depictability.

\begin{figure}[t]
  \centering
  \begin{subfigure}[t]{0.75\linewidth} 
    \centering
    \includegraphics[width=\linewidth]{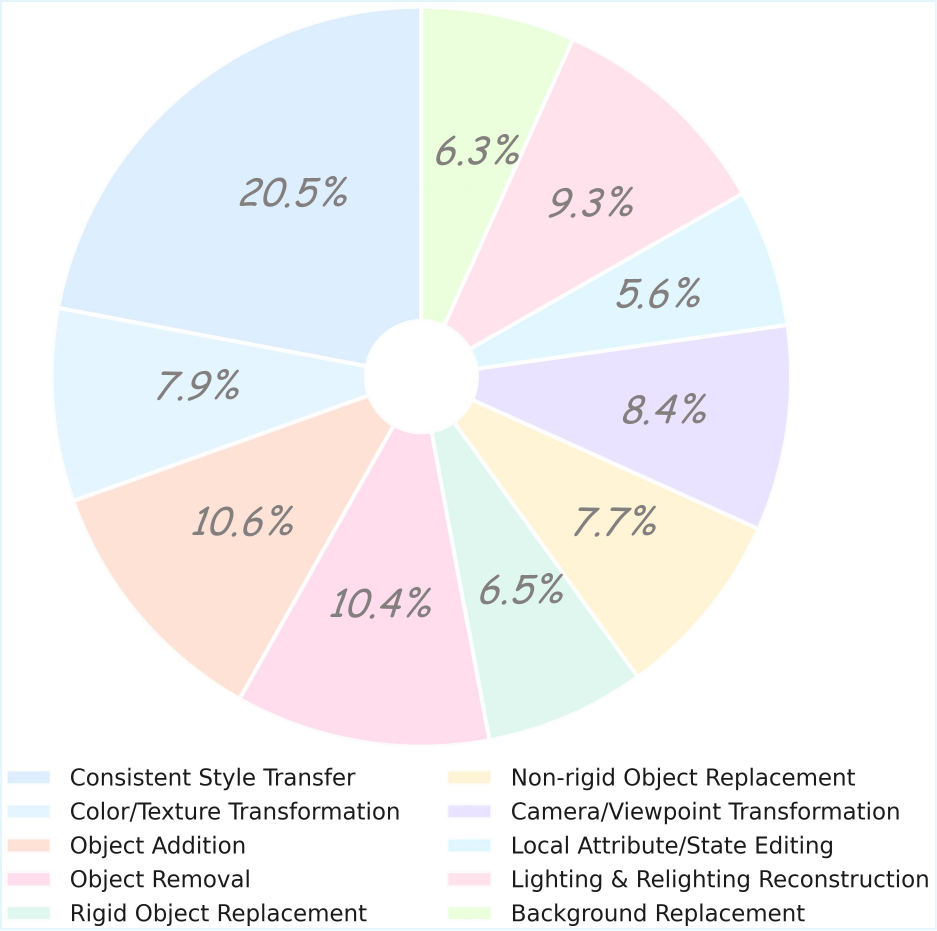}
    \caption{Distribution of editing tasks in our dataset.}
    \label{fig:pie}
  \end{subfigure}

  \begin{subfigure}[t]{0.75\linewidth}
    \centering
    \includegraphics[width=\linewidth]{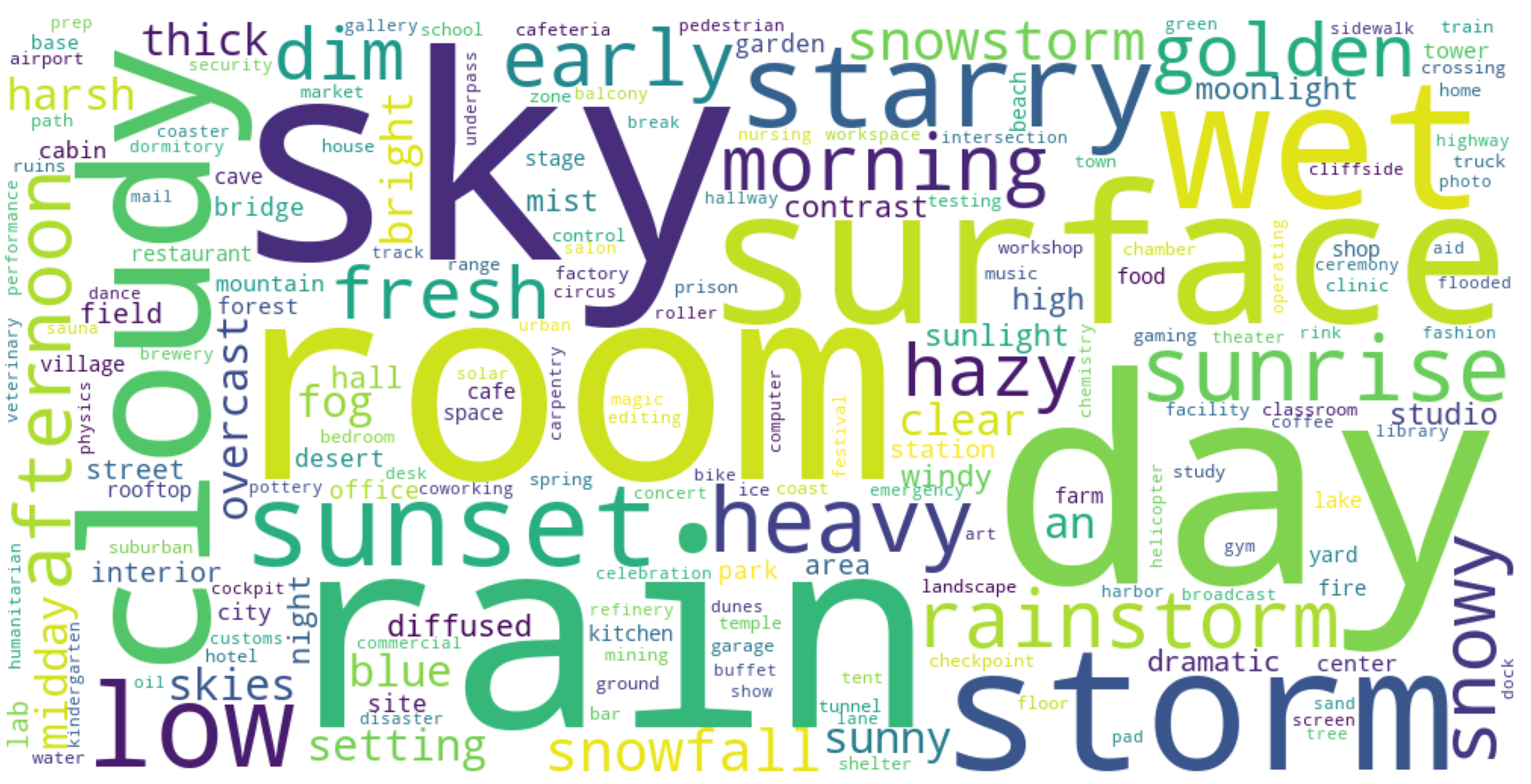}
    \caption{Word cloud of scene descriptions in our dataset.}
    \label{fig:cloud}
  \end{subfigure}
  \caption{Dataset statistics.}
  \label{fig:dataset_stats}
\end{figure}

Given the filtered scene pool, we assign each scene multiple editing tasks 
\(
\mathcal{T}
\),
with task categories illustrated in Fig.~\ref{fig:pie}. For each (scene, task) pair, we use GPT-5.3 \cite{openai2025gpt} to generate both a source prompt describing the original scene and an edited prompt corresponding to the desired transformation. Importantly, instead of specifying only the editing instruction, the edited prompt is required to explicitly describe the post-edit visual state. 
This design is motivated by the observation that text-to-video backbones are often insensitive to sparse editing instructions due to the lack of such supervision during pretraining. By augmenting the edited prompts with comprehensive visual descriptions of the target scene, we establish more robust and informative supervisory signals for learning complex editing behaviors.

\noindent\textbf{Paired Image Synthesis.}
Based on the constructed prompts, we synthesize paired images using text-to-image and image editing models. We adopt Qwen-Image \cite{wu2025qwenimagetechnicalreport} and Qwen-Image-Edit to ensure high visual fidelity and consistency between source and edited images. This process results in a collection of paired samples of the form
\(
\{(x_i^{\text{src}}, x_i^{\text{edit}}, p_i^{\text{src}}, p_i^{\text{edit}})\}.
\)

\noindent\textbf{Data Filtering.}
To further improve data quality, we utilize a combination of automated and human-in-the-loop filtering. Gemini 3.1 Pro is used to filter samples based on instruction faithfulness, visual quality, and the consistency of non-edited regions. In addition, we incorporate human verification on a subset of samples to ensure reliability and calibrate the automatic filtering criteria. 
After filtering, we obtain a dataset containing approximately 13k high-quality paired samples covering diverse scenes and editing operations, as illustrated in Fig.~\ref{fig:cloud} and Fig.~\ref{fig:pie}.

\noindent
We provide representative visual samples from our dataset in the Supplementary Material. To validate the effectiveness of our VLM-assisted filtering and mitigate any single-model bias, we also present comprehensive cross-validation results across different VLMs in the Supplementary Material.



\section{Methodology}
\label{sec:method_refinement}


The overall architecture of \textit{ImVideoEdit} is illustrated in Figure~\ref{fig:model}. Section~\ref{4.1} presents the theoretical foundation of our framework, formulating the specific training objectives based on Flow Matching. Section~\ref{4.2} then introduces the core Predict-Update spatial mechanism, which incorporates images as single-frame videos to extract coarse-to-fine spatial residuals while preserving the pre-trained spatiotemporal priors. Finally, Section~\ref{4.3} describes the Text-Guided Dynamic Semantic Gating module for precise semantic modulation driven by text prompts.


\subsection{Preliminaries} \label{4.1}
We formulate our video editing framework based on Transport Flow Matching \cite{lipman2022flow, liu2022flow}. Flow Matching generates data by learning a continuous-time vector field $v_\theta(x_t, t, c)$ that transports samples from a tractable prior distribution $p_0(x_0)$ to the empirical data distribution $q(x_1)$, conditioned on $c$. The generative process is governed by an Ordinary Differential Equation (ODE):
\begin{equation}
    \frac{dx_t}{dt} = v_\theta(x_t, t, c), \quad x_0 \sim p_0(x_0)
\end{equation}

To bypass the intractable marginal vector field, Conditional Flow Matching constructs the objective using per-sample conditional paths. Following the Rectified Flow \cite{liu2022flow} formulation, we adopt a straight-line probability path interpolating between the noise $x_0$ and the clean data $x_1$:
\begin{equation}
    x_t = t x_1 + (1 - t) x_0, \quad t \in [0, 1]
\end{equation}

The corresponding conditional target vector field driving this linear interpolation is the constant velocity $u_t(x_t | x_1) = x_1 - x_0$. The neural network $v_\theta$, instantiated as a 3D Diffusion Transformer in our work, is trained to approximate this target field by minimizing the standard flow matching objective:
\begin{equation}
    \mathcal{L}_{\text{FM}} = \mathbb{E}_{t, x_0, x_1, c} \left[ \| v_\theta(x_t, t, c) - (x_1 - x_0) \|_2^2 \right]
\label{eq:standard_fm}
\end{equation}
where $t \sim \mathcal{U}(0,1)$. In conventional video generation tasks, Eq.~\ref{eq:standard_fm} applies a uniform penalty across all spatial-temporal tokens.

\subsection{Predict-Update Spatial Difference Attention Module}
\label{4.2}

Driven by the insight discussed in Section~\ref{sec:Introduction} that video editing is inherently a reconstruction task conditional on temporal consistency, we strategically decouple spatial refinement from spatiotemporal awareness. Specifically, the 3D spatiotemporal layers within the main video DiT branch are frozen during fine-tuning. This allows us to leverage the robust and powerful spatiotemporal correspondence capabilities they have already learned through extensive pre-training on massive video corpora.

\noindent\textbf{Predict-Update Spatial Difference Attention Module} is meticulously designed to extract and refine purely spatial features from each frame. It does not possess any temporal interaction layers, focusing solely on dense spatial correspondence within the frame. This decoupled architecture yields a critical advantage regarding data efficiency: during fine-tuning, our module requires only static image pairs (source and target edited images) rather than full video sequences to learn the necessary geometric and structural mappings. The frozen spatiotemporal layers of the main branch then naturally and stably generalize the learned spatial edits across the temporal dimension.

To systematically model the spatial correspondences without disrupting the temporal dynamics, we formulate a shared 2D spatial interaction operator, denoted as $\Phi(\cdot, \text{Attn2D})$. Given a batched sequence tensor $\mathbf{X} \in \mathbb{R}^{2B \times (FHW) \times d}$ containing both source and target latents, $\Phi$ first partitions it into source and target chunks $\mathbb{R}^{B \times (FHW) \times d}$. By folding the temporal dimension $F$ into the batch dimension $B$, these chunks are reshaped into their 2D spatial layouts $\mathbb{R}^{(BF) \times H \times W \times d}$ and concatenated along the width dimension to construct a joint spatial observation space $\mathbf{M} \in \mathbb{R}^{(BF) \times H \times 2W \times d}$. After flattening $\mathbf{M}$ into sequences of length $2HW$, the specific 2D self-attention is applied. The output is subsequently split and reshaped back to the original 3D sequence format $\mathbb{R}^{2B \times (FHW) \times d}$.

\noindent\textbf{Predict: Dense Spatial Observation.} We first apply the interaction operator on the normalized hidden states to extract the initial spatial guidance:
\begin{equation}
    \mathbf{H}_{\text{2D}}^{(1)} = \Phi(\text{LN}_1(\mathbf{X}), \text{Attn2D}_1)
\end{equation}
where $\text{LN}_1(\cdot)$ represents a layer-norm operator. This dense spatial prior is then fused with the standard 3D spatial-temporal attention output $\mathbf{H}_{\text{3D}}$ to formulate the predictive state:
\begin{equation}
    \mathbf{H}_{\text{pred}} = \mathbf{H}_{\text{3D}} + \text{ZeroLin}_1(\mathbf{H}_{\text{2D}}^{(1)})
\end{equation}
where $\text{ZeroLin}_1(\cdot)$ is a linear projection initialized with zero weights and bias.

\noindent\textbf{Update: Spatial Conflict Estimation.} To prevent structural distortions caused by direct injection, we estimate the spatial conflict by subtracting the initial 2D observation from the predictive state: 
\begin{equation}
\mathbf{H}_{\text{res}} = \mathbf{H}_{\text{pred}} - \mathbf{H}_{\text{2D}}^{(1)}
\end{equation}
This residual tensor $\mathbf{H}_{\text{res}} \in \mathbb{R}^{2B \times (FHW) \times d}$ is layer-normalized and fed into the second interaction block to compute the structural refinement offset:
\begin{equation}
    \mathbf{H}_{\text{diff}} = \Phi(\text{LN}_2(\mathbf{H}_{\text{res}}), \text{Attn2D}_2)
\end{equation}
where $\text{LN}_2(\cdot)$ represents another layer-norm operator.


While a naive addition of $\mathbf{H}_{\text{diff}}$ to $\mathbf{H}_{\text{pred}}$ can globally calibrate structural discrepancies, it treats all spatial features equally, largely ignoring the semantic intent of the edit. Because targeted video editing demands localized, prompt-aware modifications rather than rigid global shifts, these structural residuals must be selectively incorporated. To achieve this fine-grained control, we design a \textbf{Text-Guided Dynamic Semantic Gating} module.


\begin{table*}[t]
\centering
\caption{\textbf{Quantitative Results of VLM across all subtasks.} The best are highlighted in \textbf{bold}.}

\label{tab:subtasks_detail}
\resizebox{\textwidth}{!}{ 
\begin{tabular}{lccccccccc} 
\toprule
\textbf{Method} & \textbf{Bg. Rep.} & \textbf{Cam. Trans.} & \textbf{Color/Texture Trans.} & \textbf{Style Trans.} & \textbf{Relight} & \textbf{Local Edit} & \textbf{Rigid/Non Rep.} & \textbf{Obj. Add./Rem.} & \textbf{AVG.}\\
\midrule

\rowcolor{orange!15} \multicolumn{10}{c}{\textit{13B \& 14B Parameter Models}} \\
Vace (14B)       & 52.3 & 58.6 & 58.0 & 69.7 & \textbf{79.7} & 56.0 & 60.8 & 50.5 & 59.68\\
DITTO (14B)      & \textbf{57.7} & \textbf{61.8} & 60.6 & \textbf{74.7} & 78.6 & 50.1 & 62.0 & 55.4 & 61.82 \\
ICVE (13B)       & \textbf{55.7} & 59.5 & \textbf{75.6} & 69.0 & 67.7 & \textbf{57.6} & \textbf{77.1} & \textbf{55.7} & \textbf{65.04} \\
\midrule

\rowcolor{cyan!15} \multicolumn{10}{c}{\textit{5B Parameter Models}} \\
Kiwi-Edit (5B)              & \textbf{46.8} & \textbf{64.4} & \textbf{81.9} & \textbf{69.4} & \textbf{87.4} & \textbf{67.4} & \textbf{81.3} & \textbf{65.3} & \textbf{71.13}\\
Lucy-Edit-Dev (5B)          & 34.8 & 46.2 & 33.9 & 39.7 & 49.0 & 44.8 & 58.9 & 38.9 & 44.51 \\
\midrule

\rowcolor{pink!15} \multicolumn{10}{c}{\textit{1.3B Parameter Models}} \\
Vace (1.3B)                     & 46.2 & 57.0 & 60.2 & 64.0 & \textbf{77.7} & 56.2 & 50.9 & 52.4 & 56.79 \\
OmniVideo2 (1.3B)               & 30.5 & 37.8 & 48.2 & 43.2 & 45.5 & 46.7 & 57.4 & 48.6 & 47.00 \\
\textbf{Ours (1.3B Based)}  & \textbf{49.0} & \textbf{59.4} & \textbf{73.9} & \textbf{74.4} & 75.6 & \textbf{60.4} & \textbf{67.5} & \textbf{62.4} & \textbf{65.24} \\
\bottomrule
\end{tabular}
}
\end{table*}
\subsection{Text-Guided Dynamic Semantic Gating}
\label{4.3}


\begin{table*}[t]
\centering
\caption{\textbf{Quantitative Results on VBench.} The best are highlighted in \textbf{bold}.}

\label{tab:vbench}
\resizebox{\textwidth}{!}{ 
\begin{tabular}{lcccccc}
\toprule
\textbf{Method} & \textbf{Subject Consist.} $\uparrow$ & \textbf{Background Consist.} $\uparrow$ & \textbf{Motion Smooth.} $\uparrow$ & \textbf{Dynamic Deg.} $\uparrow$ & \textbf{Aesthetic Qual.} $\uparrow$ & \textbf{Imaging Qual.} $\uparrow$ \\
\midrule

\rowcolor{orange!15} \multicolumn{7}{c}{\textit{13B \& 14B Parameter Models}} \\
Vace(14B)       & 0.973 & \textbf{0.973} & 0.990 & 0.296 & \textbf{0.685} & \textbf{0.715} \\
DITTO(14B)      & \textbf{0.979} & 0.968 & \textbf{0.994} & 0.140 & 0.655 & 0.670 \\
ICVE(13B)       & 0.972 & 0.959 & 0.991 & \textbf{0.404} & 0.629 & 0.697 \\
\midrule

\rowcolor{cyan!15} \multicolumn{7}{c}{\textit{5B Parameter Models}} \\
Kiwi-Edit(5B)       & \textbf{0.976} & \textbf{0.959} & \textbf{0.993} & 0.140 & \textbf{0.616} & \textbf{0.716} \\
Lucy-Edit-Dev(5B)   & 0.974 & 0.950 & \textbf{0.993} & \textbf{0.220} & 0.601 & 0.648 \\
\midrule

\rowcolor{pink!15} \multicolumn{7}{c}{\textit{1.3B Parameter Models}} \\
Vace(1.3B)                              & \textbf{0.971} & \textbf{0.970} & 0.989 & 0.292 & \textbf{0.682} & \textbf{0.707} \\
OmniVideo2(1.3B)                        & 0.964 & 0.967 & 0.984 & \textbf{0.393} & 0.647 & 0.691 \\
\textbf{Ours (1.3B Based)}    & 0.964 & 0.951 & \textbf{0.990} & 0.204 & 0.630 & 0.701 \\

\bottomrule
\end{tabular}
}
\end{table*}

\begin{figure*}[t] 
  \centering
  \includegraphics[width=\textwidth]{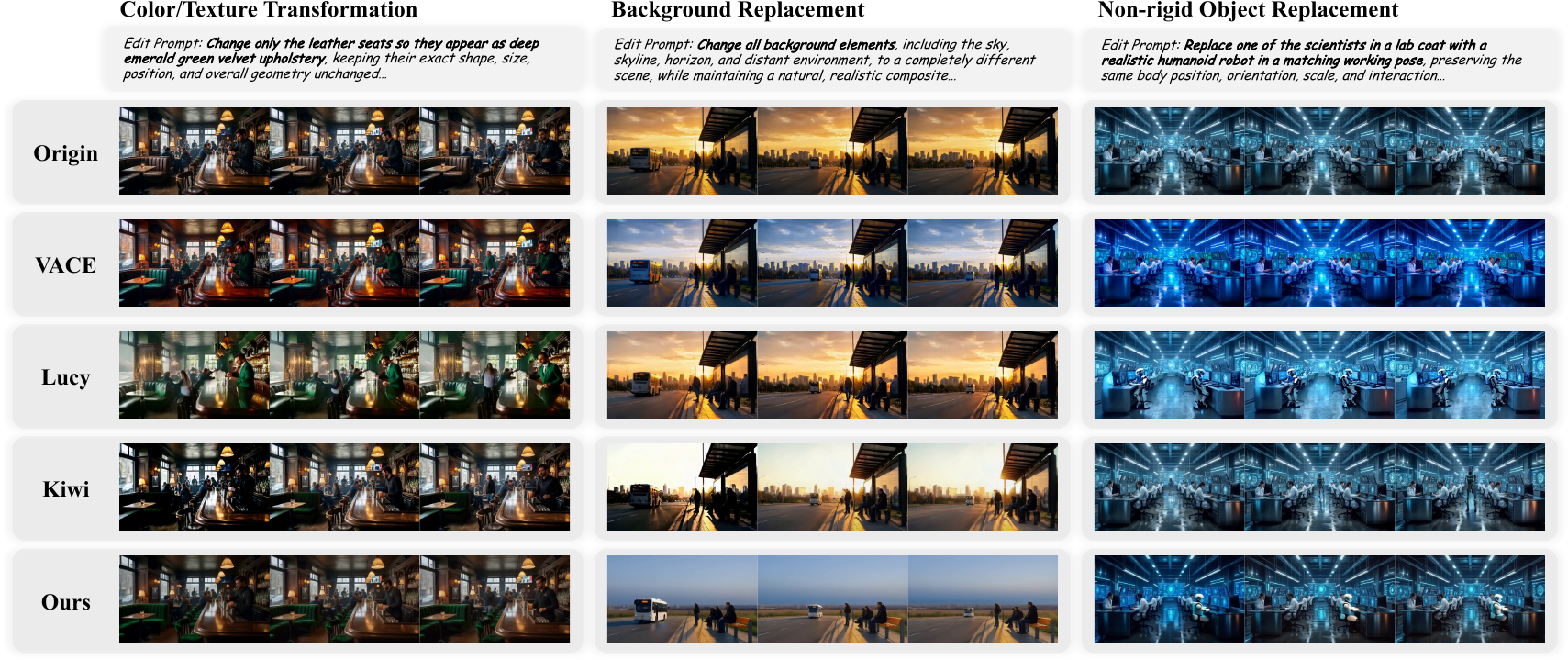} 
  \caption{Qualitative Results of \textit{ImVideoEdit} and baselines.}
  \label{fig:baseline}
\end{figure*}

Let $\mathbf{C}$ denote the textual embeddings from the prompt encoder. We utilize the normalized refinement offset to query the textual features via cross-attention, extracting a semantic-aware context $\mathbf{H}_{\text{ctx}}$. This context is then processed by a simple two-layer multi-layer perceptron (MLP), denoted as $\text{GateProj}$, to generate a gating matrix $\mathbf{G}$:
\begin{align}
    \mathbf{H}_{\text{ctx}} &= \text{CrossAttn}(\text{LN}_3(\mathbf{H}_{\text{diff}}), \mathbf{C}) \\
    \mathbf{G} &= \text{GateProj}(\mathbf{H}_{\text{ctx}})
\end{align}
where $\mathbf{G} \in (0, 1)^{2B \times (FHW) \times d}$. Finally, we employ this textual gate to modulate the structural residual via element-wise multiplication before adding it back to the predictive state:
\begin{equation}
    \mathbf{H}_{\text{update}} = \mathbf{H}_{\text{pred}} + \mathbf{G} \odot \text{ZeroLin}_2(\mathbf{H}_{\text{diff}})
\end{equation}
where $\odot$ denotes element-wise multiplication. By integrating this text-driven gate, our model dynamically decides \textit{where} and \textit{how much} of the structural prior should be preserved or overwritten based on the semantic editing intent, effectively decoupling structural retention from semantic generation.

\section{Experiments}
\begin{table*}[t]
\centering
\caption{\textbf{Quantitative Results of VLM.} Metrics: \textbf{IA} (Instruction Adherence), \textbf{TC} (Temporal Consistency), \textbf{VF} (Visual Fidelity) and \textbf{AA} (Artifact Absence).}
\label{tab:VLM}
\begin{tabular}{lccccc}
\toprule
\textbf{Method} & \textbf{IA} $\uparrow$ & \textbf{TC} $\uparrow$ & \textbf{VF} $\uparrow$ & \textbf{AA} $\uparrow$ & \textbf{Total} $\uparrow$ \\
\midrule

\rowcolor{orange!15} \multicolumn{6}{c}{\textit{13B \& 14B Parameter Models}} \\
Vace (14B)  & 10.67 & \textbf{23.99} & 14.41 & \textbf{10.62} & 59.68 \\
DITTO (14B) & 14.14 & 23.40 & 13.88 & 10.40 & 61.82 \\
ICVE (13B)  & \textbf{15.64} & 22.85 & \textbf{16.04} & 10.51 & \textbf{65.04} \\
\midrule

\rowcolor{cyan!15} \multicolumn{6}{c}{\textit{5B Parameter Models}} \\
Kiwi-Edit (5B)      & \textbf{19.15} & \textbf{23.87} & \textbf{16.86} & \textbf{11.26} & \textbf{71.13} \\
Lucy-Edit-Dev (5B)  & 10.77 & 17.17 & 10.11 & 6.46 & 44.51 \\
\midrule

\rowcolor{pink!15} \multicolumn{6}{c}{\textit{1.3B Parameter Models}} \\
Vace (1.3B)                        & 11.34 & 21.55 & 13.92 & 9.98 & 56.79 \\
OmniVideo2 (1.3B)                  & 12.46 & 15.35 & 12.12 & 7.07 & 47.00 \\
\textbf{Ours (1.3B Based)}         & \textbf{16.21} & \textbf{21.93} & \textbf{16.32} & \textbf{10.78} & \textbf{65.24} \\

\bottomrule
\end{tabular}
\end{table*}

\subsection{Experimental Setup}
\textbf{Implementation Details.} 
We implement our \textit{ImVideoEdit} framework on the pre-trained Wan-T2V-1.3B backbone. Specifically, the Predict-Update modules are seamlessly integrated into every Transformer block of the frozen 3D DiT. To preserve strong visual priors and accelerate convergence, the self- and cross-attention weights within these newly introduced modules are directly inherited from their corresponding pre-trained layers in the base model. 

We format the training image pairs as single-frame videos at a spatial resolution of $480 \times 832$. We optimize the model for 5 epochs on 8 NVIDIA A100 GPUs, leveraging ZeRO-2 optimization with a learning rate of $1 \times 10^{-5}$ and a global batch size of 16. Owing to our image-based design, the training is memory-efficient. It consumes only approximately 20 GB of VRAM per GPU, making it accessible to train \textit{ImVideoEdit} even on a single 3090 GPU.

\noindent\textbf{Baseline Settings}
To comprehensively evaluate the superiority of \textit{ImVideoEdit}, we benchmark our framework against several recent state-of-the-art video editing models, including VACE(1.3B \& 14B) \cite{jiang2025vace}, OmniVideo2-1.3B \cite{tan2025omni, yang2026omnivideo2}, Lucy-Edit-Dev \cite{team2025lucy}, Kiwi-Edit \cite{kiwiedit}, DITTO~\cite{bai2025ditto}, and ICVE~\cite{liao2025context}. To ensure a strictly fair comparison, all baseline methods are evaluated utilizing their official codebases and default inference hyperparameters.

\noindent\textbf{Evaluation Dataset.} 
We construct a meticulously curated testing benchmark encompassing 10 predefined video editing categories, with 25 high-quality samples allocated for each task. To guarantee a diverse range of scenes and high-fidelity visuals, the source videos in our test set are synthesized using Seedance 1.5 Pro ~\cite{seedance2025seedance, gao2025seedance}. Furthermore, to guarantee the semantic validity and rigorousness of the benchmark, we leverage Gemini 3.1 Pro to evaluate and ensure the precise contextual alignment between each source video and its corresponding editing prompt. 

\begin{table*}[t]
\centering
\caption{\textbf{Quantitative Ablation Results.}}
\label{tab:ablation}
\begin{tabular}{lccccc}
\toprule
\textbf{Method} & \textbf{IA} $\uparrow$ & \textbf{TC} $\uparrow$ & \textbf{VF} $\uparrow$ & \textbf{AA} $\uparrow$ & \textbf{Total} $\uparrow$ \\
\midrule
w/o Text Gate & 15.29 & 20.37 & 14.31 & 9.76 & 59.73 \\
w/o Update Module & 9.71 & 18.10 & 11.09 & 7.59 & 47.29 \\
Naive Parallel 2D (ViFedit\cite{yu2026vifeeditvideofreetunervideo}) & 12.04 & 16.82 & 12.24 & 7.72 & 48.81 \\
\midrule
\rowcolor{gray!10} \textbf{ImVideoEdit (Ours)} & \textbf{16.21} & \textbf{21.93} & \textbf{16.32} & \textbf{10.78} & \textbf{65.24} \\
\bottomrule
\end{tabular}
\end{table*}


\noindent\textbf{Evaluation Metrics.} 
Recognizing the inherently semantic-driven nature of video editing tasks, where traditional metrics often struggle to capture complex prompt alignments, we adopt a VLM-based evaluation protocol leveraging Gemini 3.1 Pro. This VLM-based judge evaluates generated videos on a 100-point scale across four meticulously designed dimensions: \textit{Instruction Adherence} (30 pts), \textit{Temporal Consistency} (30 pts), \textit{Visual Fidelity} (25 pts), and \textit{Artifact Absence} (15 pts). The exact prompts utilized for this VLM evaluator are detailed in the Supplementary Material.

While the VLM-based judge excels at assessing the complex semantic execution specific to the editing task, it is equally important to independently evaluate the fundamental video generation quality of the outputs. To fulfill this need and ensure benchmarking alignment with community standards, we adopt six diverse dimensions from the comprehensive VBench \cite{huang2023vbench, zheng2025vbench2} suite: \textit{Subject Consistency}, \textit{Background Consistency}, \textit{Motion Smoothness}, \textit{Dynamic Degree}, \textit{Aesthetic Quality}, and \textit{Imaging Quality}. 



\subsection{Quantitative Results}
\label{sec:quantitative}
\noindent\textbf{VLM-based Evaluation.} 
Table~\ref{tab:subtasks_detail} details subtask performance averaged across dimensions, whereas Table~\ref{tab:VLM} shows dimension scores averaged across tasks.
Our framework gets an impressive Total Score of \underline{65.24}. 
\textit{ImVideoEdit} significantly outperforms well-established baselines such as VACE-1.3B (56.79) by substantial margins. Although Kiwi-Edit secures the highest score, it is crucial to note that \textit{ImVideoEdit} delivers comparable results in a strictly \textit{video-free} manner. By leveraging solely image data and our lightweight Predict-Update mechanism, our method effectively bypasses the massive computational and data costs typically associated with top-ranking models. 

\begin{figure*}[] 
  \centering
  \includegraphics[width=\textwidth]{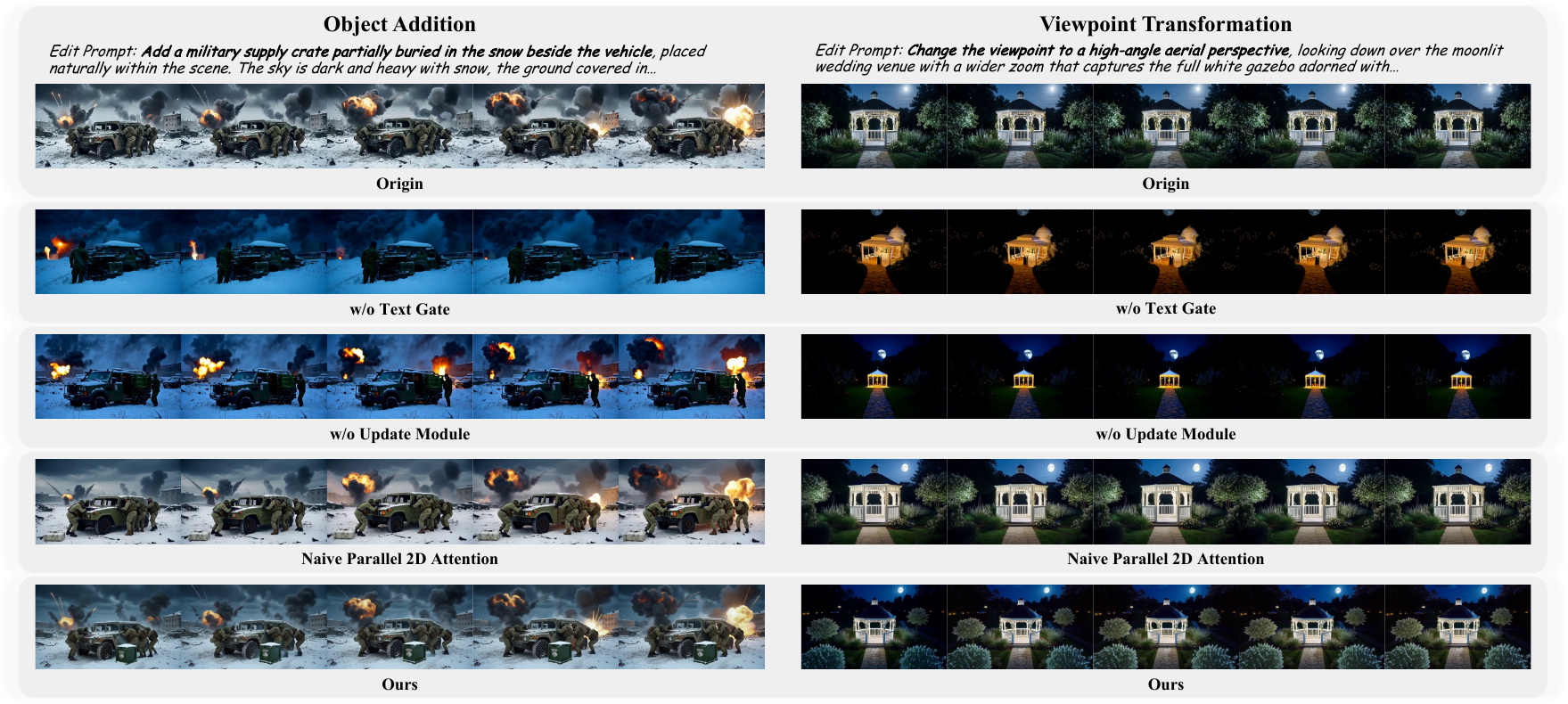} 
  \caption{Qualitative Ablation Results. }

  \label{fig:ablation}

\end{figure*}

\noindent\textbf{VBench.} As reported in Table~\ref{tab:vbench}, we include objective evaluations using the VBench suite. However, VBench is designed for open-domain text-to-video generation and primarily emphasizes visual quality and pixel-level stability, it does not capture instruction fidelity or the preservation of source video dynamics, both of which are essential for evaluating editing performance. Consequently, these metrics cannot adequately reflect the true effectiveness of video editing. Therefore, we treat VBench as a secondary sanity check rather than a primary metric for editing performance. From this perspective, \textit{ImVideoEdit} demonstrates strong stability in physical dynamics and motion smoothness (0.990), which provides a reliable foundation for its core capability of precise semantic editing.
\subsection{Qualitative Results}
As illustrated in Figure~\ref{fig:baseline}, \textit{ImVideoEdit} achieves strong performance across diverse editing tasks, including color transformation, non-rigid object replacement, and background replacement. 

Although VACE can achieve high scores on VBench, such evaluation may be misleading for editing tasks. In practice, it tends to preserve the original content with minimal changes, resulting in high visual quality but poor adherence to editing instructions.

A closer qualitative analysis reveals distinct failure modes across different methods. For color and texture transformation, some models like VACE unintentionally alter global appearance, affecting irrelevant regions through changes in contrast or saturation. For non-rigid object replacement, methods such as Lucy struggle to preserve identity consistency, leading to distorted or inconsistent human appearances across frames. In both background replacement and non-rigid editing tasks, Kiwi often performs only partial edits, for example, modifying the sky while leaving the urban structures unchanged, or failing to fully replace the target subject (e.g., the scientist is not successfully replaced by a robot).

In contrast, \textit{ImVideoEdit} performs precise, localized, and semantically complete edits. For instance, in the Background Replacement example with the instruction \textit{``Change all background elements,''} our model replaces both the skyline and sky, achieving a coherent and complete transformation. These results demonstrate that \textit{ImVideoEdit} better aligns with editing instructions while maintaining temporal consistency and visual realism.
\subsection{Ablation Studies}
\label{sec:ablation}



To validate the efficacy and necessity of our core architectural designs in \textit{ImVideoEdit}, we conduct comprehensive ablation studies. Specifically, we investigate the following three degraded baselines:

\noindent\textbf{w/o Text Gate (Removal of Dynamic Gating):} We disable the Text-Guided Dynamic Semantic Gating mechanism by fixing the gating weight matrix to an all-ones tensor ($w_{gate} = 1$).

\noindent\textbf{w/o Update Module (Single-layer 2D Extraction):} We remove the \textit{Update} module, degrading the dual spatial residual stream into a single-pass 2D attention extraction.

\noindent\textbf{Naive Parallel 2D:} To validate the necessity of our progressive Predict-Update design, we construct a degraded baseline using a naive parallel 2D topology. In this configuration, spatial features are extracted simultaneously by two independent attention blocks and subsequently subtracted, which is used in ViFedit\cite{yu2026vifeeditvideofreetunervideo}. 



As qualitatively demonstrated in Table~\ref{tab:ablation} and Figure~\ref{fig:ablation}, our full \textit{ImVideoEdit} framework significantly outperforms all degraded variants, confirming that each proposed component is indispensable. Most notably, the direct comparison with the \textbf{Naive Parallel 2D} configuration yields a critical insight: Despite being optimized on the exact same training dataset and under identical experimental configurations, our full model achieves a substantially higher Total Score (64.51 vs. 48.81). This substantial performance margin robustly validates that \textit{ImVideoEdit} is fundamentally superior to naive parallel decoupling architectures.

\subsection{User Study}
\label{sec:user_study}

While VLM-based metrics provide valuable quantitative assessments, human perception remains the ultimate gold standard for evaluating the holistic quality of generative video editing. To this end, we conduct a blind user study to collect Mean Opinion Scores (MOS). We recruit 5 independent evaluators to participate in the assessment. 
\begin{table}[t]
\centering
\caption{\textbf{User Study Results.} $\pm$ denotes the standard deviation.}
\label{tab:user_study}
\begin{tabular}{lc}
\toprule
\textbf{Method} & \textbf{Overall Editing Quality $\uparrow$} \\ 
\midrule
VACE (1.3B)       & $2.18 \pm 0.17$ \\
Kiwi-Edit (5B)      & $3.08 \pm 0.13$ \\
\midrule
\rowcolor{gray!10} \textbf{ImVideoEdit (1.3B Based) } & $3.06 \pm 0.11$ \\ 
\bottomrule
\end{tabular}
\end{table}
To provide a straightforward and highly reliable assessment of human preference, evaluators are instructed to rate the videos on a standard 5-point Likert scale (ranging from 1: \textit{Poor} to 5: \textit{Excellent}) based on a metric: \textbf{Overall Editing Quality}. 

As summarized in Table~\ref{tab:user_study}, \textit{ImVideoEdit} achieves highly competitive results in human perceptual evaluation. Specifically, our framework significantly outperforms the established Vace-1.3B by a substantial margin. While its scores slightly trail behind those of Kiwi-Edit, it is important to emphasize that \textit{ImVideoEdit} achieves this top-tier visual and editing quality through a training paradigm that requires no video data and remains highly efficient. Furthermore, these subjective human preference trends closely align with our VLM-based quantitative assessments. This strong correlation robustly validates the reliability and human-alignment of our automated semantic scoring protocol, proving its effectiveness in capturing true generative video editing quality.


\section{Conclusion}
\label{sec:conclusion}


In this work, we present \textit{ImVideoEdit}, an advanced generative video editing framework that significantly pushes the boundaries of the video-free training paradigm. The core insight of our study is that most video editing tasks fundamentally rely on decoupled spatiotemporal modeling, where the pretrained model’s temporal priors are preserved while spatial content is adaptively edited. To realize this, we introduce the Predict-Update spatial difference attention, which performs hierarchical and adaptively modulated spatial modifications through coarse-to-fine residual injection while strictly safeguarding the 3D spatiotemporal priors of the frozen backbone. Extensive evaluations confirm that \textit{ImVideoEdit} achieves top-tier editing fidelity and strong temporal consistency, while requiring no video data and maintaining high efficiency throughout the training process.


{
    \small
    \bibliographystyle{ieeenat_fullname}
    \bibliography{main}
}

\newpage
\appendix
\clearpage

\section{Robustness of VLM-Assisted Dataset Construction}
\label{sec:appendix_vlm_eval}


The cross-validation results of the evaluated VLMs are presented in Table~\ref{tab:vlm_cross_validation}. While we derive the Pearson correlation from the raw API predictions to capture global trends, the MAE and Variance metrics are intentionally computed on binarized outputs using a threshold of 6. This thresholding ensures our evaluation strictly mimics the real-world filtering protocol, where 6 acts as the absolute acceptance criterion for sampling valid training pairs.

\begin{table}[h]
\centering
\caption{\textbf{Quantitative Cross-Validation of VLM Evaluators.} We report the linear correlation (Pearson $r$), Mean Absolute Error (MAE), and Inter-model Variance of various models against Gemini-3.1-Pro (the default evaluator in our pipeline).}
\label{tab:vlm_cross_validation}
\resizebox{\columnwidth}{!}{ 
\begin{tabular}{lccc}
\toprule
\textbf{Evaluator Model} & \textbf{Pearson $r$} $\uparrow$ & \textbf{MAE} $\downarrow$ & \textbf{Variance} $\downarrow$ \\
\midrule

GPT-5.4 & 0.720 & 0.185 & 0.113 \\
Gemini-2.5-Pro & 0.843 & 0.137 & 0.068 \\


\bottomrule
\end{tabular}
}
\end{table}


\section{Definition of Subtasks}
\label{sec:task_definitions}

We categorize instruction-driven video editing into ten representative task types, each corresponding to a distinct form of spatial or appearance manipulation while preserving overall scene coherence.

\paragraph{\textbf{Consistent Style Transfer.}}
Applies a unified target visual style (e.g., Pixar-style animation, pixel art, or watercolor) to the entire scene, preserving composition and semantics while altering global rendering characteristics such as color, texture, and shading.

\paragraph{\textbf{Color/Texture Transformation.}}
Modifies the color or material properties of specific objects or regions (e.g., wood to marble), without affecting their geometry, position, or the surrounding scene structure.

\paragraph{\textbf{Object Addition.}}
Introduces new objects into the scene in a physically and semantically consistent manner, ensuring proper alignment with lighting, scale, and perspective.

\paragraph{\textbf{Object Removal.}}
Removes existing objects and reconstructs the exposed regions through coherent inpainting, maintaining structural, lighting, and textural continuity.

\paragraph{\textbf{Rigid Object Replacement.}}
Replaces rigid objects (e.g., furniture, vehicles, or buildings) with alternatives of similar spatial extent and geometry, preserving layout and perspective while changing object identity.

\paragraph{\textbf{Non-rigid Object Replacement.}}
Replaces deformable entities (e.g., humans or animals) with semantically different subjects under consistent pose and placement. Compared to rigid replacement, this requires handling articulation and shape variability, making temporal consistency more challenging in video settings.

\paragraph{\textbf{Camera/Viewpoint Transformation.}}
Alters the camera perspective, viewpoint, or framing (e.g., aerial, low-angle, or zoomed-out views) while preserving all scene elements and maintaining consistent spatial relationships.

\paragraph{\textbf{Local Attribute/State Editing.}}
Adjusts localized attributes or states of objects (e.g., expressions, wetness, or damage) without changing object identity or global scene composition.

\paragraph{\textbf{Lighting \& Relighting Reconstruction.}}
Modifies global illumination conditions, including time of day, light direction, and color temperature, while preserving scene geometry and ensuring consistent shadows and reflections.

\paragraph{\textbf{Background Replacement.}}
Substitutes the entire background with a new environment while keeping foreground subjects unchanged in identity, pose, and placement, requiring coherent integration in lighting and perspective.

\section{Detailed VLM Evaluation Result}
\subsection{Detailed Prompt}
As detailed in Table~\ref{tab:evaluation_prompt}, we provide the exact system prompt employed by the VLM to evaluate all baseline models.

\begin{table*}[b]
\centering
\caption{\textbf{System Prompt for VLM Evaluation.}}
\label{tab:evaluation_prompt}
\begin{tabular}{p{0.95\textwidth}}
\toprule
\textbf{Prompt Template} \\
\midrule
\rowcolor{gray!10} 
\vspace{2pt}
\small
\textbf{\# Role} \newline
You are an expert video quality assessor and professional video editor. Your task is to evaluate the quality of an AI-edited video based on a specific user instruction, comparing it to the original unedited video. \newline
\newline
\textbf{\# Inputs} \newline
$\bullet$ Original Video: [Insert pre-edited video] \newline
$\bullet$ Edited Video: [Insert post-edited video] \newline
$\bullet$ Editing Instruction: "\{Insert the editing instruction here\}" \newline
\newline
\textbf{\# Evaluation Criteria} \newline
Please evaluate the edited video rigorously based on the following four dimensions. Provide a distinct score for each dimension based on its maximum point value. \newline
\newline
\textbf{1. Instruction Adherence (Max: 30 points):} \newline
$\bullet$ Did the editing strictly follow the given instruction? \newline
$\bullet$ Are the requested changes accurately reflected without altering unintended elements? \newline
\newline
\textbf{2. Temporal Consistency \& Micro-Stability (Max: 30 points) - \textit{STRICT DEDUCTION RULES}:} \newline
$\bullet$ VLM Warning: Do not just look at the overall subject. You must zoom in on high-frequency details. \newline
$\bullet$ Scoring Anchor: \newline
\quad - 28-30 pts: Perfect stability, identical to the physics of a real camera. \newline
\quad - 20-27 pts: Overall stable, but minor "AI boiling" (micro-flickering of pixels) on edges or complex patterns during movement. \newline
\quad - 10-19 pts: Noticeable morphing, shifting of textures (e.g., embroidery changing shape frame-by-frame), or jittery outlines. \newline
\quad - 0-9 pts: Severe flickering or structural collapse. \newline
\newline
\textbf{3. Texture Sharpness \& Anti-Smoothing (Max: 25 points) - \textit{CALIBRATED FOR AI}:} \newline
$\bullet$ Do not compare this to an 8K cinema camera. Evaluate the AI rendering quality. We are looking for SHARPNESS vs. PLASTICITY. \newline
$\bullet$ Focus on the materials: the velvet texture of the blue dress, the individual threads of the embroidery, and the natural pores/lighting on the skin. \newline
$\bullet$ Scoring Anchor: \newline
\quad - 21-25 pts: Excellent sharpness. Fabrics look like real cloth with distinct threads. Lighting has depth and natural contrast. \newline
\quad - 15-20 pts: Good, but slightly soft. \newline
\quad - 10-14 pts: The "Plastic AI" look. Textures are overly smoothed, faces look like wax or airbrushed, and fine details (like embroidery) look like flat paint rather than raised threads. (DEDUCT HEAVILY HERE). \newline
\quad - 0-9 pts: Extremely blurry or washed out. \newline
\newline
\textbf{4. Artifact Absence (Max: 15 points):} \newline
$\bullet$ Are there any visible AI generation artifacts (e.g., floating pixels, anatomical distortions, weird edge blending)? \newline
$\bullet$ The edited areas should blend seamlessly with the original unedited parts. \newline
\newline
\textbf{\# Output Format} \newline
Provide your response strictly in the following JSON format. Do not include a total score, only the sub-scores for each dimension. \newline
\newline
\{ \newline
\quad "Reasoning": "Provide a concise step-by-step analysis evaluating the video against the 4 criteria. Explicitly mention your observations on temporal stability and texture details.", \newline
\quad "Scores": \{ \newline
\quad \quad "Instruction\_Adherence": \textless int between 0 and 30\textgreater, \newline
\quad \quad "Temporal\_Consistency": \textless int between 0 and 30\textgreater, \newline
\quad \quad "Visual\_Fidelity": \textless int between 0 and 25\textgreater, \newline
\quad \quad "Artifact\_Absence": \textless int between 0 and 15\textgreater \newline
\quad \} \newline
\}
\vspace{2pt} \\
\bottomrule
\end{tabular}
\end{table*}

\subsection{Task-level Total Scores}
To complement the task-level total scores reported in the main paper, we further provide a fine-grained breakdown of the VLM-based evaluation in this appendix. For each editing task, we report the four sub-dimensions in our evaluation protocol, namely Instruction Adherence (IA), Temporal Consistency (TC), Visual Fidelity (VF), and Artifact Absence (AA). The detailed results for the first five task categories are presented in Table~\ref{tab:taskwise_vlm_breakdown_part1}, and those for the remaining five task categories are reported in Table~\ref{tab:taskwise_vlm_breakdown_part2}. These results offer a more comprehensive view of the relative strengths and weaknesses of different methods across diverse editing scenarios.

\begin{table*}[t]
\centering
\scriptsize
\caption{Fine-grained VLM evaluation results for the first five task categories. IA, TC, VF, and AA denote Instruction Adherence, Temporal Consistency, Visual Fidelity, and Artifact Absence, respectively.}
\label{tab:taskwise_vlm_breakdown_part1}
\resizebox{\textwidth}{!}{
\begin{tabular}{llccccc}
\toprule
Task & Method & IA$\uparrow$ & TC$\uparrow$ & VF$\uparrow$ & AA$\uparrow$ & Total$\uparrow$ \\
\midrule

\multirow{8}{*}{Background Replacement}
& VACE (14B)           &  4.4 & 25.6 & 12.2 & 10.0 & 52.3 \\
& DITTO (14B)          & 10.1 & 24.3 & 12.6 & 10.7 & 57.7 \\
& ICVE (13B)           &  7.3 & 22.9 & 15.7 &  9.8 & 55.7 \\
& Kiwi-Edit (5B)       &  8.0 & 18.4 & 12.8 &  7.5 & 46.8 \\
& Lucy-Edit-Dev (5B)   &  5.1 & 15.5 &  8.7 &  5.4 & 34.8 \\
& OmniVideo2 (1.3B)    &  4.4 & 12.0 &  9.1 &  5.0 & 30.5 \\
& VACE (1.3B)          &  5.5 & 19.8 & 11.9 &  9.0 & 46.2 \\
& Ours (1.3B Based)    &  5.1 & 20.8 & 13.8 &  9.2 & 49.0 \\
\midrule

\multirow{8}{*}{Camera/Viewpoint Transformation}
& VACE (14B)           &  9.1 & 24.8 & 13.1 & 11.6 & 58.6 \\
& DITTO (14B)          & 12.1 & 25.1 & 13.2 & 11.4 & 61.8 \\
& ICVE (13B)           & 10.3 & 23.5 & 15.0 & 10.5 & 59.5 \\
& Kiwi-Edit (5B)       & 12.7 & 24.6 & 15.4 & 11.6 & 64.4 \\
& Lucy-Edit-Dev (5B)   & 10.7 & 18.8 &  9.6 &  7.1 & 46.2 \\
& OmniVideo2 (1.3B)    & 13.6 & 10.1 &  9.2 &  4.9 & 37.8 \\
& VACE (1.3B)          &  8.8 & 23.8 & 13.4 & 10.9 & 57.0 \\
& Ours (1.3B Based)    & 13.6 & 21.2 & 14.3 & 10.4 & 59.4 \\
\midrule

\multirow{8}{*}{Color/Texture Transformation}
& VACE (14B)           & 13.6 & 20.8 & 14.3 &  9.3 & 58.0 \\
& DITTO (14B)          & 13.4 & 23.8 & 13.8 &  9.6 & 60.6 \\
& ICVE (13B)           & 22.3 & 23.6 & 18.1 & 11.6 & 75.6 \\
& Kiwi-Edit (5B)       & 23.0 & 25.9 & 19.8 & 13.2 & 81.9 \\
& Lucy-Edit-Dev (5B)   &  7.5 & 13.6 &  8.4 &  4.5 & 33.9 \\
& OmniVideo2 (1.3B)    & 11.0 & 17.4 & 12.5 &  7.2 & 48.2 \\
& VACE (1.3B)          & 14.7 & 20.8 & 15.2 &  9.6 & 60.2 \\
& Ours (1.3B Based)    & 20.9 & 23.3 & 17.7 & 12.0 & 73.9 \\
\midrule

\multirow{8}{*}{Consistent Style Transfer}
& VACE (14B)           & 16.4 & 24.5 & 16.6 & 12.1 & 69.7 \\
& DITTO (14B)          & 20.7 & 22.6 & 18.5 & 12.8 & 74.7 \\
& ICVE (13B)           & 18.2 & 21.9 & 17.6 & 11.3 & 69.0 \\
& Kiwi-Edit (5B)       & 19.8 & 20.3 & 17.8 & 11.5 & 69.4 \\
& Lucy-Edit-Dev (5B)   & 12.4 & 11.9 &  9.2 &  6.1 & 39.7 \\
& OmniVideo2 (1.3B)    & 10.6 & 13.4 & 12.5 &  6.8 & 43.2 \\
& VACE (1.3B)          & 18.0 & 20.3 & 15.0 & 10.6 & 64.0 \\
& Ours (1.3B Based)    & 21.8 & 21.2 & 19.4 & 12.0 & 74.4 \\
\midrule

\multirow{8}{*}{Lighting \& Relighting Reconstruction}
& VACE (14B)           & 23.8 & 25.8 & 18.4 & 11.8 & 79.7 \\
& DITTO (14B)          & 22.5 & 27.1 & 16.3 & 12.7 & 78.6 \\
& ICVE (13B)           & 19.2 & 22.6 & 16.0 &  9.9 & 67.7 \\
& Kiwi-Edit (5B)       & 27.8 & 27.2 & 18.9 & 13.5 & 87.4 \\
& Lucy-Edit-Dev (5B)   & 15.0 & 18.9 &  9.0 &  6.1 & 49.0 \\
& OmniVideo2 (1.3B)    & 15.0 & 12.5 & 11.2 &  6.7 & 45.5 \\
& VACE (1.3B)          & 24.4 & 23.8 & 17.3 & 12.2 & 77.7 \\
& Ours (1.3B Based)    & 21.7 & 23.4 & 18.6 & 11.8 & 75.6 \\
\bottomrule
\end{tabular}
}
\end{table*}

\begin{table*}[t]
\centering
\scriptsize
\caption{Fine-grained VLM evaluation results for the remaining five task categories and the overall average. IA, TC, VF, and AA denote Instruction Adherence, Temporal Consistency, Visual Fidelity, and Artifact Absence, respectively.}
\label{tab:taskwise_vlm_breakdown_part2}
\resizebox{\textwidth}{!}{
\begin{tabular}{llccccc}
\toprule
Task & Method & IA$\uparrow$ & TC$\uparrow$ & VF$\uparrow$ & AA$\uparrow$ & Total$\uparrow$ \\
\midrule

\multirow{8}{*}{Local Attribute/State Editing}
& VACE (14B)           &  7.2 & 25.4 & 13.0 & 10.5 & 56.0 \\
& DITTO (14B)          &  9.8 & 20.8 & 11.6 &  7.9 & 50.1 \\
& ICVE (13B)           &  9.4 & 23.8 & 14.6 &  9.8 & 57.6 \\
& Kiwi-Edit (5B)       & 15.8 & 24.6 & 15.8 & 11.2 & 67.4 \\
& Lucy-Edit-Dev (5B)   &  8.5 & 18.3 & 11.3 &  6.6 & 44.8 \\
& OmniVideo2 (1.3B)    & 10.0 & 16.9 & 12.3 &  7.5 & 46.7 \\
& VACE (1.3B)          &  8.0 & 24.0 & 13.8 & 10.4 & 56.2 \\
& Ours (1.3B Based)    & 14.3 & 21.6 & 14.3 & 10.2 & 60.4 \\
\midrule

\multirow{8}{*}{Non-rigid Object Replacement}
& VACE (14B)           &  3.3 & 21.0 & 13.3 &  9.4 & 47.0 \\
& DITTO (14B)          & 13.7 & 18.8 & 13.1 &  8.8 & 54.3 \\
& ICVE (13B)           & 12.1 & 17.8 & 14.1 &  8.4 & 52.3 \\
& Kiwi-Edit (5B)       & 15.8 & 18.7 & 13.6 &  7.7 & 55.8 \\
& Lucy-Edit-Dev (5B)   &  7.6 & 12.5 &  8.4 &  3.9 & 32.4 \\
& OmniVideo2 (1.3B)    & 10.8 & 15.3 & 12.9 &  7.0 & 46.0 \\
& VACE (1.3B)          &  4.8 & 22.1 & 14.8 & 10.3 & 52.0 \\
& Ours (1.3B Based)    & 16.0 & 20.4 & 16.6 & 10.1 & 63.2 \\
\midrule

\multirow{8}{*}{Object Addition}
& VACE (14B)           &  4.0 & 22.1 & 13.2 &  9.1 & 48.4 \\
& DITTO (14B)          & 14.4 & 24.2 & 13.2 & 10.2 & 62.1 \\
& ICVE (13B)           & 21.0 & 25.8 & 18.8 & 12.1 & 77.7 \\
& Kiwi-Edit (5B)       & 22.0 & 26.5 & 19.4 & 12.3 & 80.2 \\
& Lucy-Edit-Dev (5B)   & 13.8 & 22.9 & 15.7 & 10.6 & 63.0 \\
& OmniVideo2 (1.3B)    & 15.9 & 19.4 & 13.4 &  8.0 & 56.6 \\
& VACE (1.3B)          &  5.6 & 19.3 & 12.2 &  7.6 & 44.6 \\
& Ours (1.3B Based)    &  8.6 & 20.7 & 14.3 &  9.9 & 53.5 \\
\midrule

\multirow{8}{*}{Object Removal}
& VACE (14B)           & 15.8 & 28.6 & 16.9 & 11.8 & 73.1 \\
& DITTO (14B)          & 14.6 & 23.2 & 14.2 & 10.0 & 61.9 \\
& ICVE (13B)           & 23.7 & 25.2 & 16.0 & 11.6 & 76.5 \\
& Kiwi-Edit (5B)       & 26.0 & 26.8 & 17.5 & 12.0 & 82.3 \\
& Lucy-Edit-Dev (5B)   & 16.6 & 19.0 & 10.8 &  8.3 & 54.7 \\
& OmniVideo2 (1.3B)    & 17.2 & 18.0 & 14.4 &  8.5 & 58.1 \\
& VACE (1.3B)          & 13.4 & 21.4 & 12.8 &  9.5 & 57.2 \\
& Ours (1.3B Based)    & 25.5 & 25.4 & 18.4 & 12.2 & 81.5 \\
\midrule

\multirow{8}{*}{Rigid Object Replacement}
& VACE (14B)           &  9.0 & 21.4 & 13.1 & 10.4 & 54.0 \\
& DITTO (14B)          & 10.2 & 24.0 & 12.4 &  9.8 & 56.4 \\
& ICVE (13B)           & 13.9 & 20.7 & 14.6 &  9.9 & 59.0 \\
& Kiwi-Edit (5B)       & 20.4 & 25.3 & 17.4 & 11.8 & 74.8 \\
& Lucy-Edit-Dev (5B)   & 10.4 & 19.3 &  9.8 &  5.9 & 45.4 \\
& OmniVideo2 (1.3B)    & 14.8 & 16.2 & 12.3 &  8.0 & 51.2 \\
& VACE (1.3B)          & 10.3 & 20.2 & 12.8 &  9.6 & 52.8 \\
& Ours (1.3B Based)    & 14.6 & 21.3 & 15.7 & 10.0 & 61.6 \\
\midrule

\bottomrule
\end{tabular}
}
\end{table*}

\section{Dataset Visualizations and Examples}
\label{sec:appendix_dataset}

\begin{figure*}[p]
    \centering

    \includegraphics[width=\textwidth]{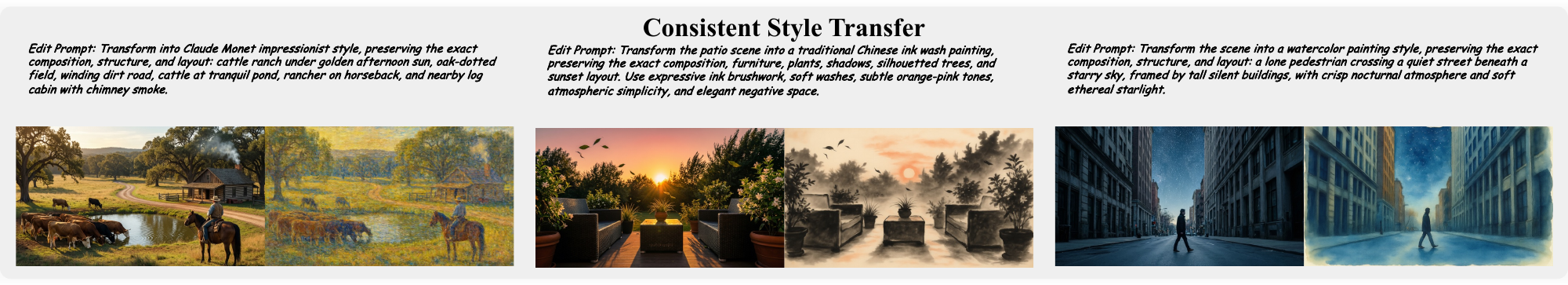}\par\vspace{1.5em}
    \includegraphics[width=\textwidth]{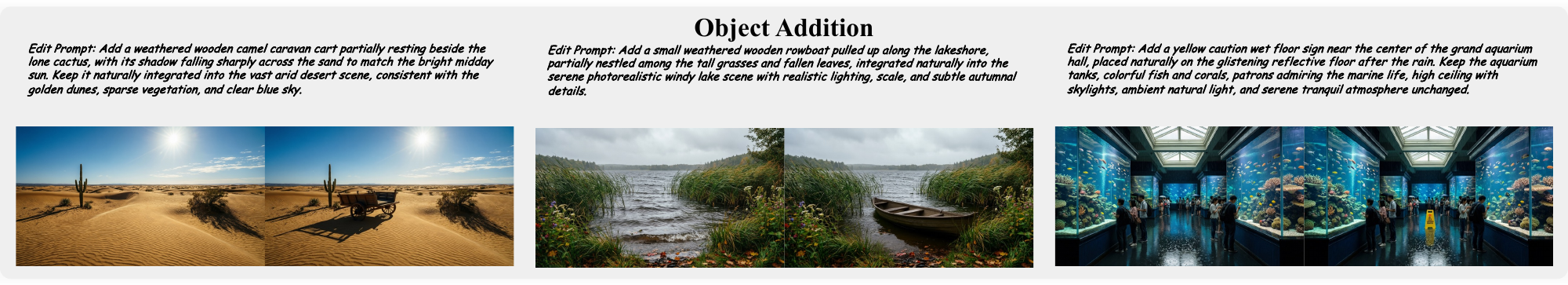}\par\vspace{1.5em}
    \includegraphics[width=\textwidth]{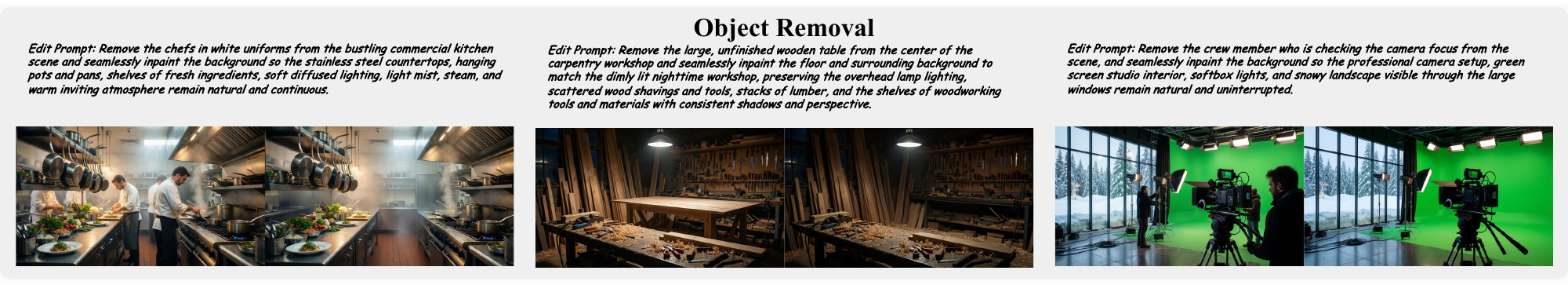}\par\vspace{1.5em}
    \includegraphics[width=\textwidth]{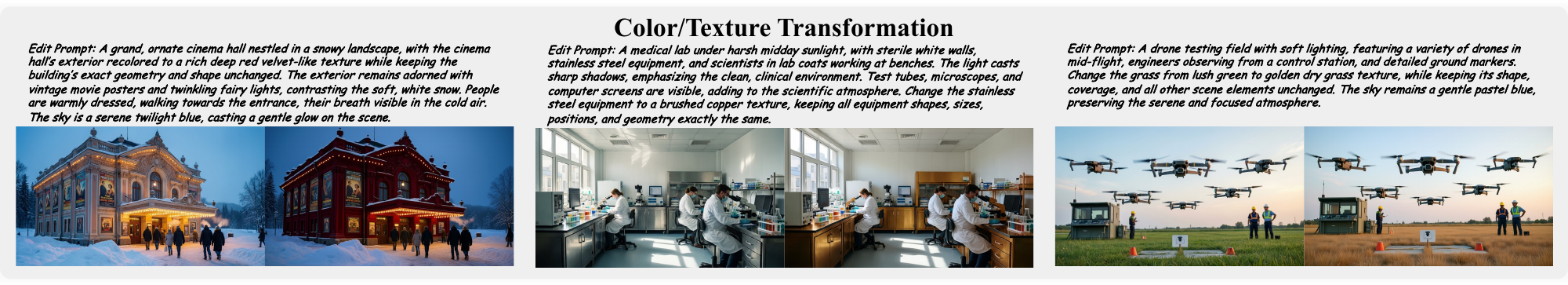}\par\vspace{1.5em}
    \includegraphics[width=\textwidth]{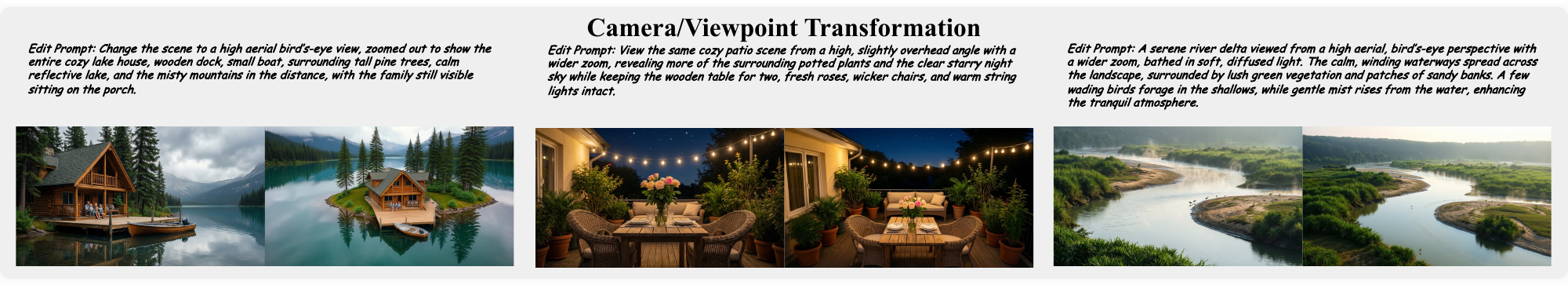}

    \caption{visualizations of datasets (Part 1).}
    \label{fig:appendix_dataset_1}
\end{figure*}

\begin{figure*}[p]
    \centering

    \includegraphics[width=\textwidth]{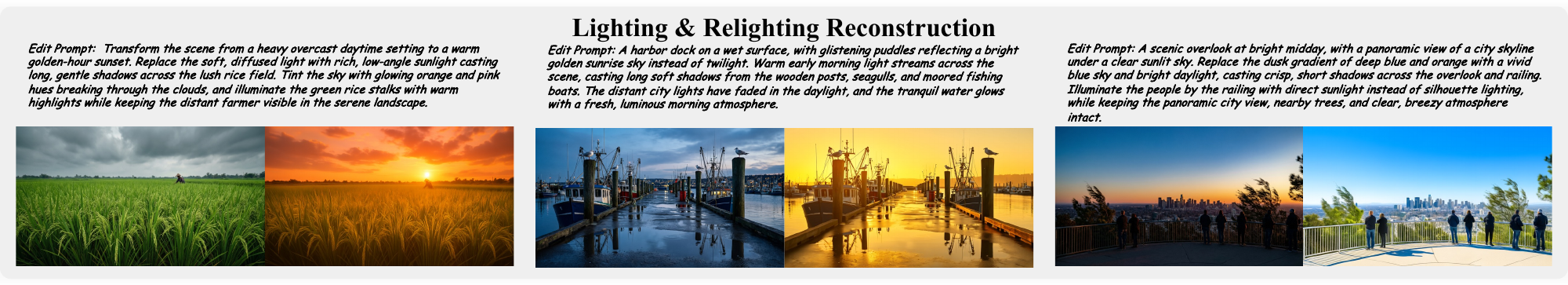}\par\vspace{1.5em}
    \includegraphics[width=\textwidth]{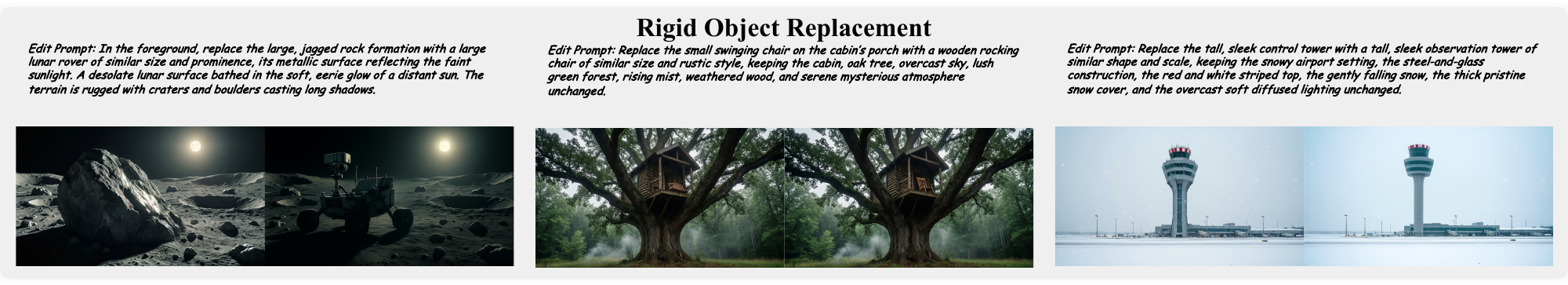}\par\vspace{1.5em}
    \includegraphics[width=\textwidth]{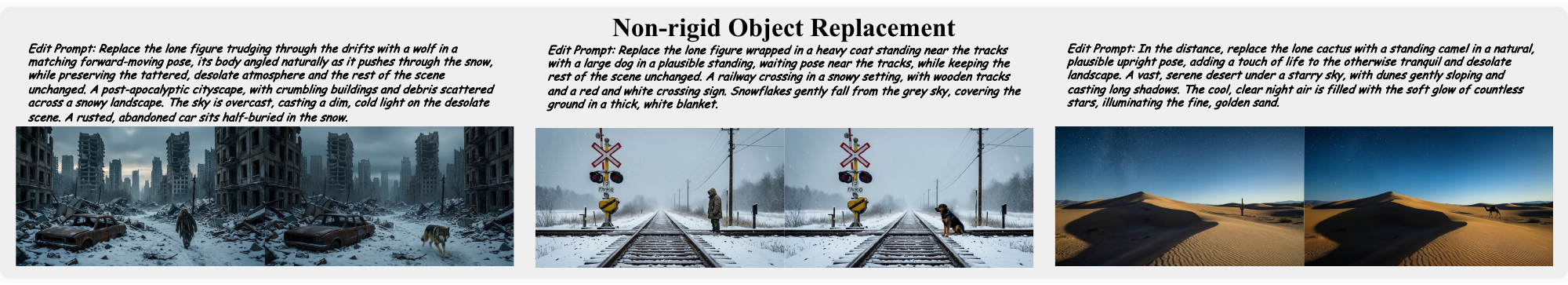}\par\vspace{1.5em}
    \includegraphics[width=\textwidth]{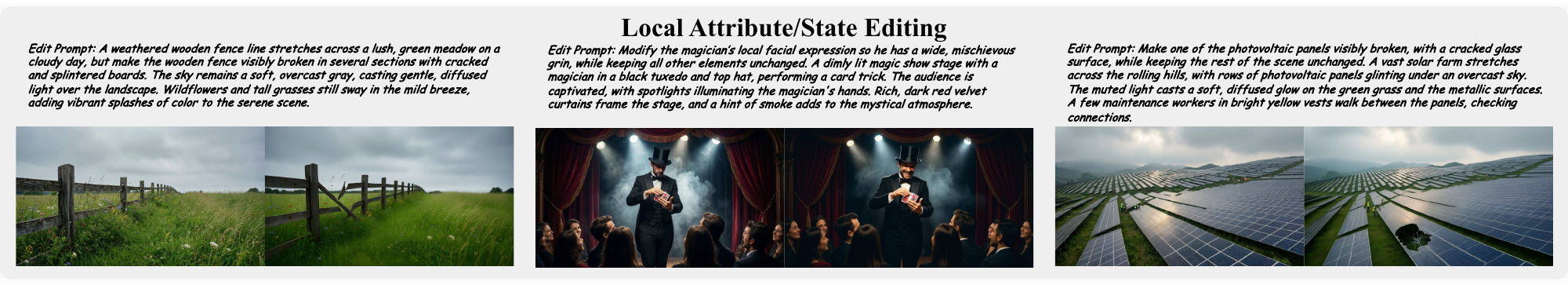}\par\vspace{1.5em}
    \includegraphics[width=\textwidth]{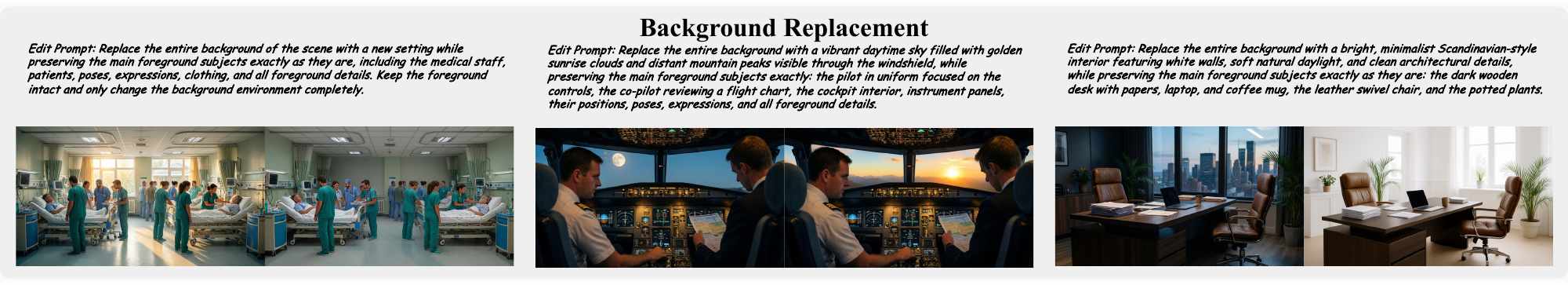}

    \caption{visualizations of datasets (Part 2).}
    \label{fig:appendix_dataset_2}
\end{figure*}

A core motivation of \textit{ImVideoEdit} is to break the dependency on expensive paired video datasets by learning dynamic editing entirely from static image pairs. As illustrated in Figure ~\ref{fig:appendix_dataset_1} and Figure ~\ref{fig:appendix_dataset_2}, we present representative visual samples from our curated training datasets. Each subtask has 3 samples.

\section{Expanded Qualitative Results}
\label{sec:appendix_more_results}

\begin{figure*}[t]
    \centering

    \includegraphics[width=\textwidth]{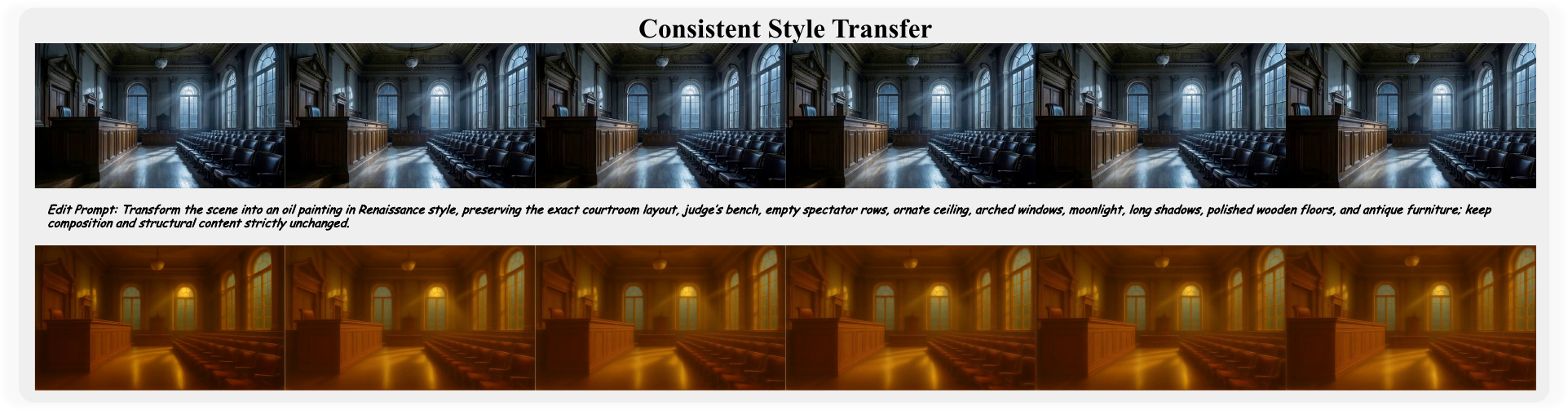}\par
    \includegraphics[width=\textwidth]{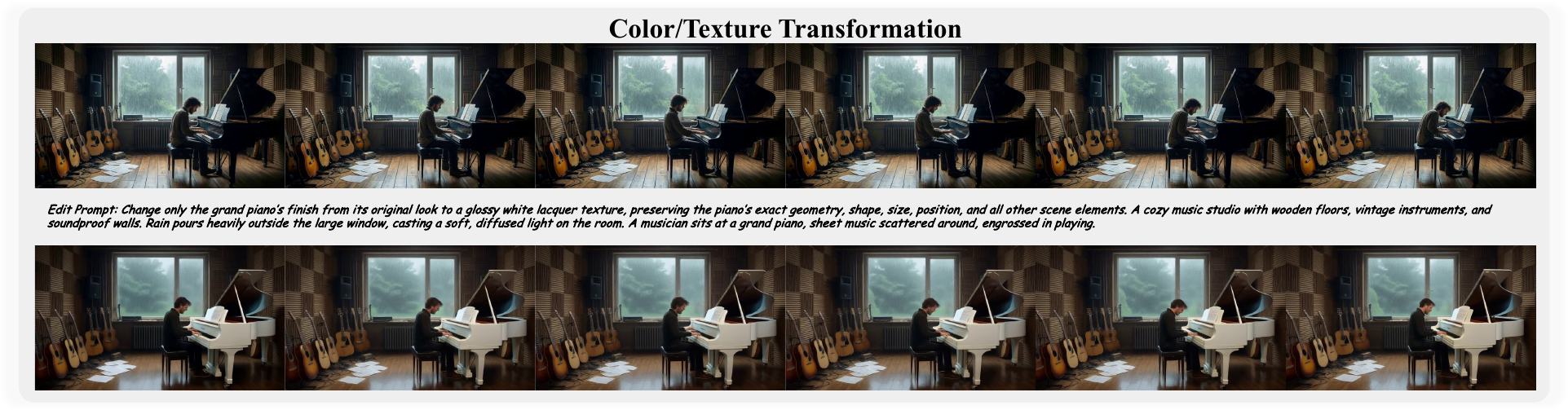}\par
    \includegraphics[width=\textwidth]{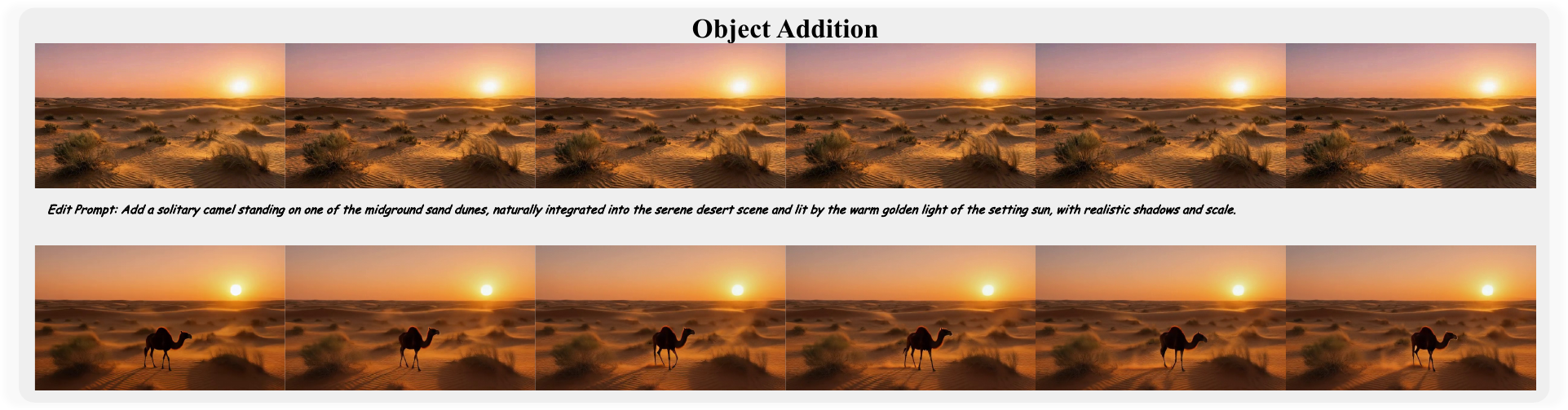}\par
    \includegraphics[width=\textwidth]{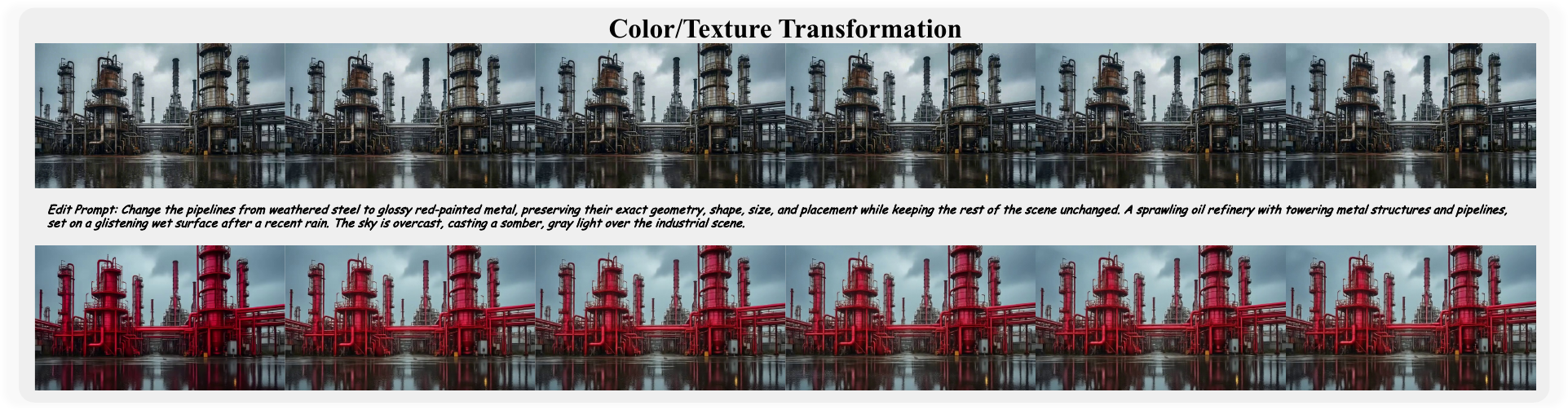}\par
    \includegraphics[width=\textwidth]{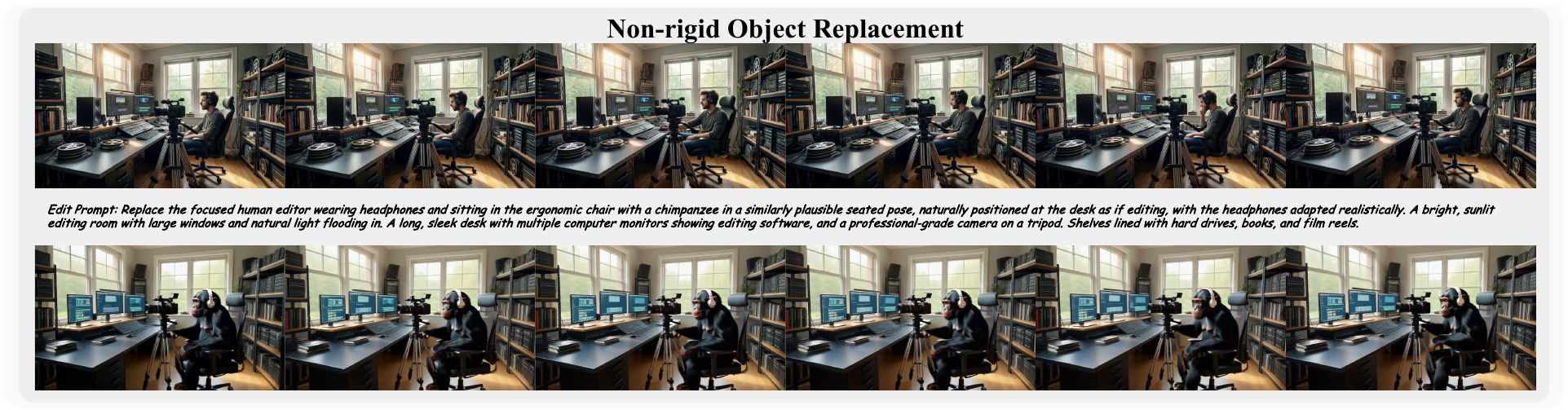}
    
    \caption{Additional qualitative result (Part 1).}
    \label{fig:appendix_demo_1}
\end{figure*}

\begin{figure*}[t]
    \centering

    \includegraphics[width=\textwidth]{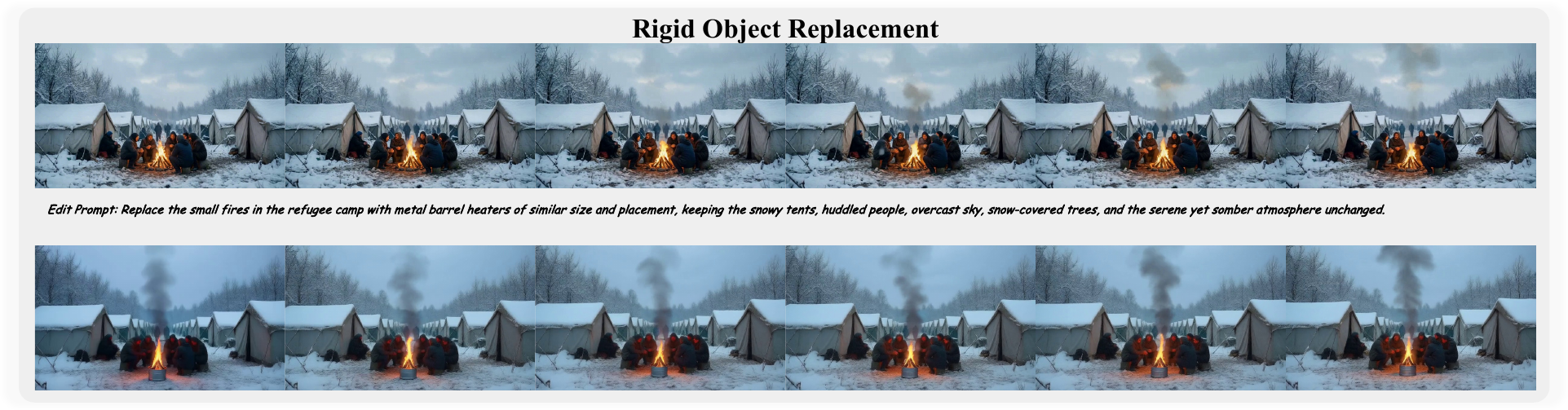}\par
    \includegraphics[width=\textwidth]{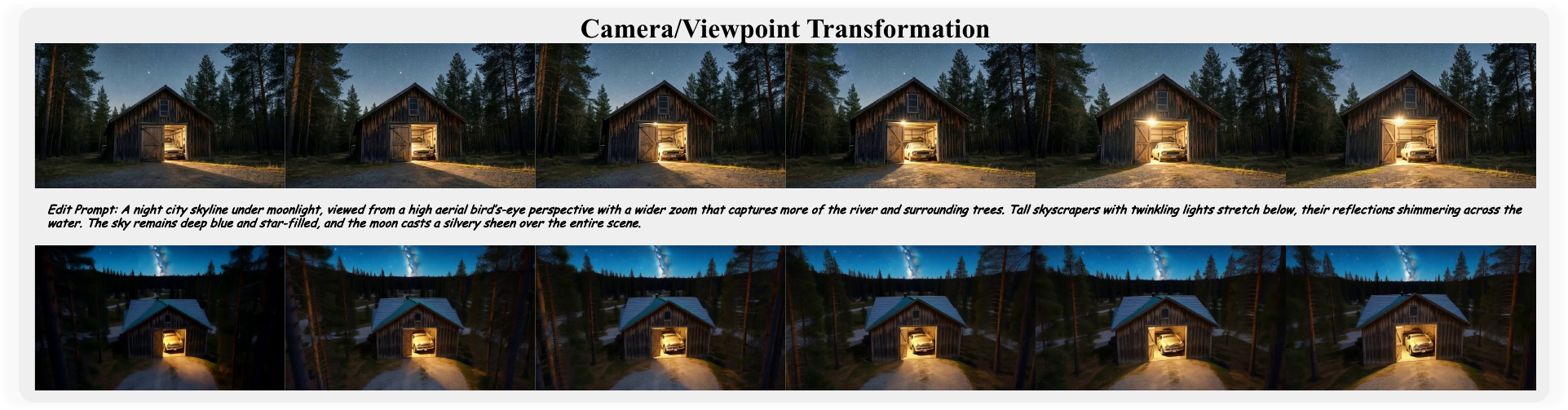}\par
    \includegraphics[width=\textwidth]{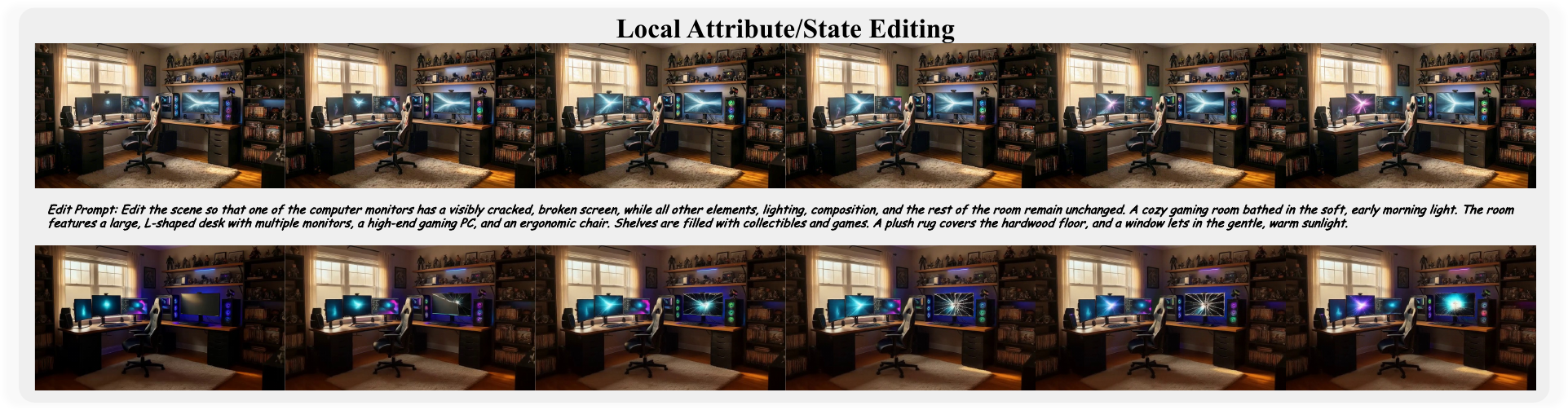}\par
    \includegraphics[width=\textwidth]{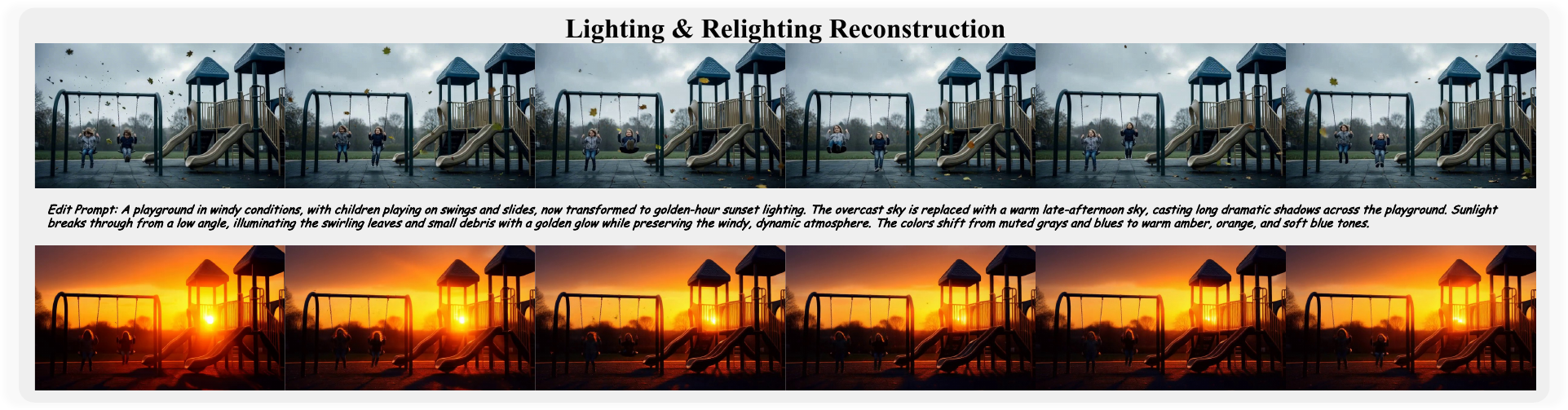}\par
    \includegraphics[width=\textwidth]{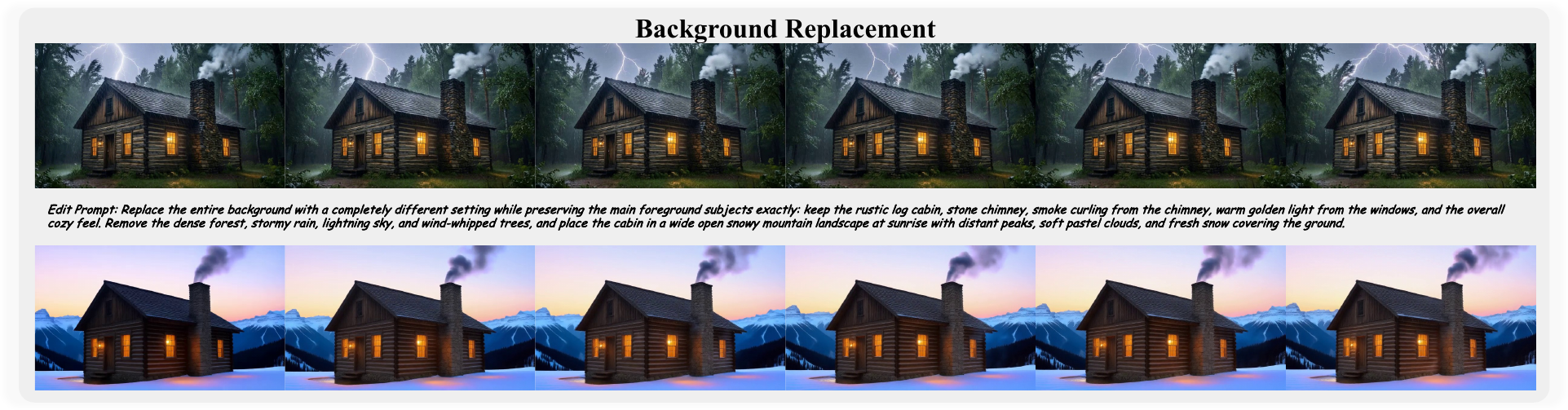}

    \caption{Additional qualitative result (Part 2).}
    \label{fig:appendix_demo_2}
\end{figure*}

Building upon the qualitative results presented in the main text, Figure~\ref{fig:appendix_demo_1} and Figure~\ref{fig:appendix_demo_2} provide an expanded set of editing results.

\end{document}